\documentclass[10pt,twocolumn,letterpaper]{article}

\usepackage{cvpr}              

\usepackage[accsupp]{axessibility}
\usepackage{graphicx}
\usepackage{amsmath}
\usepackage{amssymb}
\usepackage{booktabs}

\usepackage{times}
\usepackage{epsfig}

\usepackage{url}
\usepackage{amsthm}
\usepackage{algorithm}
\usepackage{algorithmic}
\usepackage{multirow}
\usepackage{amsfonts} 
\usepackage{bbding}
\usepackage{makecell}
\usepackage{colortbl}
\usepackage{xcolor}
\usepackage{microtype}
\usepackage{bm}
\usepackage{wrapfig}
\newtheorem{myDef}{Definition}
\newtheorem{myTheo}{Theorem}
\newtheorem{myAxio}{Axioms}

\makeatletter
\newcommand{\ssymbol}[1]{$^{\@fnsymbol{#1}}$}
\makeatother

\newcommand{\tabfootnotesize}{\fontsize{8}{9}\selectfont}

\newcommand{\myparagraph}[1]{\textbf{#1}\hspace{1.8ex}}

\definecolor{cred}{HTML}{FF6B6B}
\definecolor{corange}{HTML}{FF5200}
\definecolor{cgreen}{HTML}{70AD47}
\definecolor{cblue}{HTML}{686EE2}
\definecolor{cpurple}{HTML}{A149FA}
\definecolor{ggray}{RGB}{127,127,127}
\definecolor{aliceblue}{rgb}{0.94, 0.97, 1.0}

\usepackage[pagebackref,breaklinks,colorlinks]{hyperref}

\usepackage[capitalize]{cleveref}
\crefname{section}{Sec.}{Secs.}
\Crefname{section}{Section}{Sections}
\Crefname{table}{Table}{Tables}
\crefname{table}{Tab.}{Tabs.}

\def\cvprPaperID{4337} 
\def\confName{CVPR}
\def\confYear{2023}

\begin{document}


\title{Video-Text as Game Players:\\ Hierarchical Banzhaf Interaction for Cross-Modal Representation Learning}

\author{%
    Peng Jin$^{1,3}$ \quad
    Jinfa Huang$^{1,3}$ \quad
    Pengfei Xiong$^{4}$ \quad
    Shangxuan Tian$^{4}$ \quad
    Chang Liu$^{5}$ \quad
    \and
    Xiangyang Ji$^{5}$ \quad
    Li Yuan$^{1,2,3}$\footnotemark[1] \quad
    Jie Chen$^{1,2,3}$\footnotemark[1]\\[3pt]
    \small{$^1$School of Electronic and Computer Engineering, Peking University, Shenzhen, China} \quad
    \small{$^2$Peng Cheng Laboratory, Shenzhen, China} \\
    \small{$^3$AI for Science (AI4S)-Preferred Program, Peking University Shenzhen Graduate School, Shenzhen, China} \\
    \small{$^4$Shopee, Shenzhen, China} \quad \small{$^5$Department of Automation and BNRist, Tsinghua University, Beijing, China} \\
    \footnotesize{\{jp21, jinfahuang\}@stu.pku.edu.cn} \quad
    \footnotesize{xiongpengfei@gmail.com} \quad \footnotesize{tianshangxuan@u.nus.edu} \\ \footnotesize{\{liuchang2022, xyji\}@tsinghua.edu.cn} \quad
    \footnotesize{yuanli-ece@pku.edu.cn \quad chenj@pcl.ac.cn}
}

\maketitle

\renewcommand{\thefootnote}{\fnsymbol{footnote}}

\footnotetext[1]{Corresponding author: Li Yuan, Jie Chen.}

\begin{abstract}
Contrastive learning-based video-language representation learning approaches, e.g., CLIP, have achieved outstanding performance, which pursue semantic interaction upon pre-defined video-text pairs. To clarify this coarse-grained global interaction and move a step further, we have to encounter challenging shell-breaking interactions for fine-grained cross-modal learning. In this paper, we creatively model video-text as game players with multivariate cooperative game theory to wisely handle the uncertainty during fine-grained semantic interaction with diverse granularity, flexible combination, and vague intensity. Concretely, we propose \underline{H}ierarchical \underline{B}anzhaf \underline{I}nteraction (HBI) to value possible correspondence between video frames and text words for sensitive and explainable cross-modal contrast. To efficiently realize the cooperative game of multiple video frames and multiple text words, the proposed method clusters the original video frames (text words) and computes the Banzhaf Interaction between the merged tokens. By stacking token merge modules, we achieve cooperative games at different semantic levels. Extensive experiments on commonly used text-video retrieval and video-question answering benchmarks with superior performances justify the efficacy of our HBI. More encouragingly, it can also serve as a visualization tool to promote the understanding of cross-modal interaction, which have a far-reaching impact on the community. Project page is available at \href{https://jpthu17.github.io/HBI/}{https://jpthu17.github.io/HBI/}.
\end{abstract}

\section{Introduction}
Representation learning based on both vision and language has many potential benefits and direct applicability to cross-modal tasks, such as text-video retrieval~\cite{jin2022expectationmaximization,luo2021clip4clip} and video-question answering~\cite{xu2017video,li2022representation}. Visual-language learning has recently boomed due to the success of contrastive learning~\cite{wu2018unsupervised,chen2020a,he2020momentum,chen2021an,chen2020improved,zhang2023patchlevel,zhang2021zero,zhang2022m,zhang2022align}, \eg, CLIP~\cite{radford2021learning}, to project the video and text features into a common latent space according to the semantic similarities of video-text pairs. In this manner, cross-modal contrastive learning enables networks to learn discriminative video-language representations.

\begin{figure}[tbp]
\centering
\includegraphics[width=.95\linewidth]{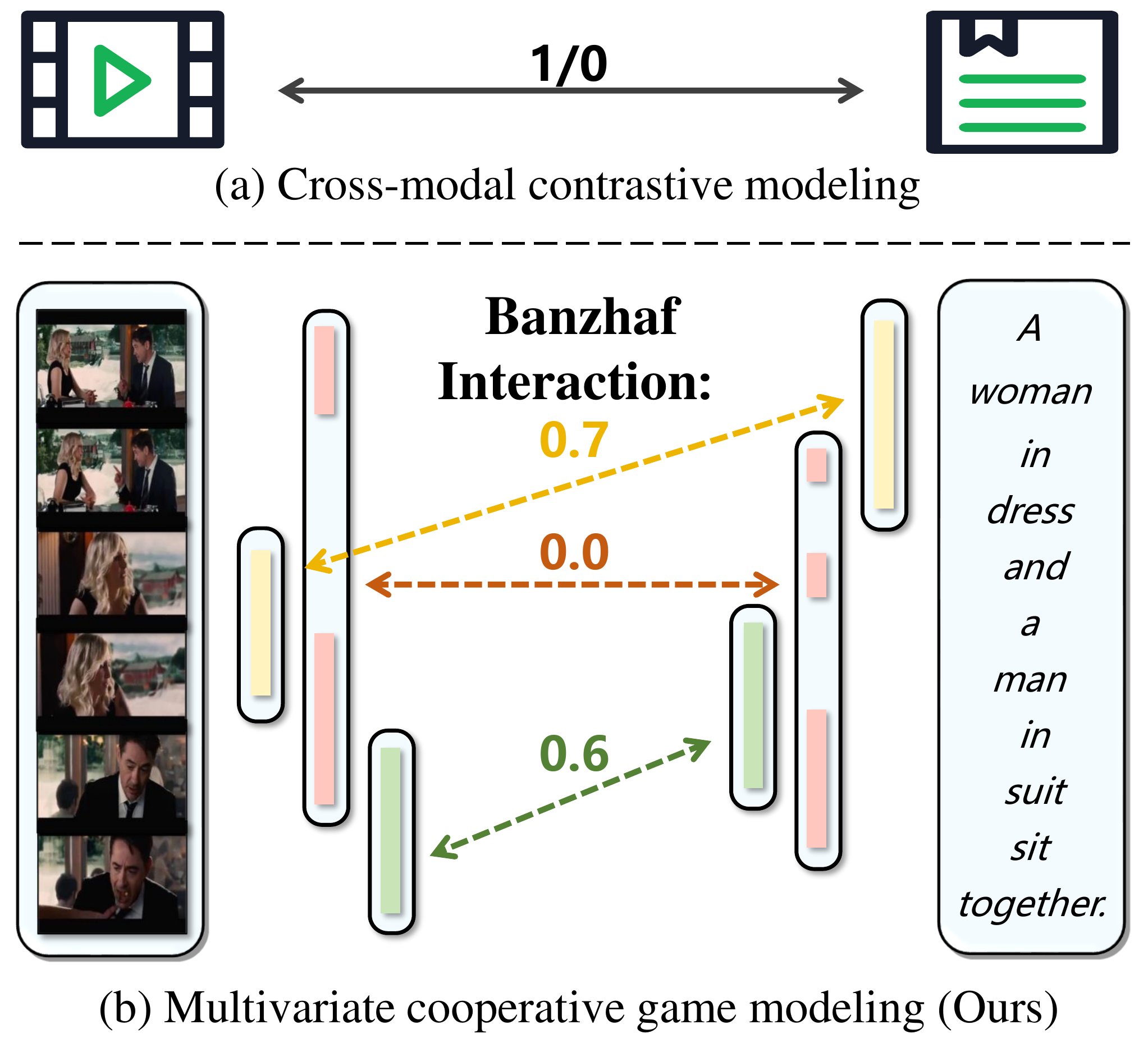}
\vspace{-0.5em}
\caption{(a) Cross-modal contrastive methods only learn a global semantic interaction from the coarse-grained labels of video-text pairs. (b) We model cross-modal alignment as a multivariate cooperative game process. Specifically, we use Banzhaf Interaction to value possible correspondence between video frames and text words and consider it as an additional learning signal.}
\label{fig0}
\vspace{-.8em}
\end{figure}

The cross-modal contrastive approach~\cite{luo2021clip4clip,dong2021dual,jin2022expectationmaximization} typically models the cross-modal interaction via solely the global similarity of each modality. Specifically, as shown in Fig.~\ref{fig0}a, it only exploits the coarse-grained labels of video-text pairs to learn a global semantic interaction. However, in most cases, we expect to capture fine-grained interpretable information, such as how much cross-modal alignment is helped or hindered by the interaction of a visual entity and a textual phrase. Representation that relies on cross-modal contrastive learning cannot do this in a supervised manner, as manually labeling these interpretable relationships is unavailable, especially on large-scale datasets. This suggests that there might be other learning signals that could complement and improve pure contrastive formulations.

In contrast to prior works~\cite{luo2021clip4clip,wang2022disentangled,jin2022expectationmaximization}, we model cross-modal representation learning as a multivariate cooperative game by formulating video and text as players in a cooperative game, as illustrated in Fig.~\ref{fig0}b. Intuitively, if visual representations and textual representations have strong semantic correspondence, they tend to cooperate together and contribute to the cross-modal similarity score. Motivated by this spirit, we consider the set containing multiple representations as a coalition, and propose to quantify the trend of cooperation within a coalition via the game-theoretic interaction index, \ie, Banzhaf Interaction~\cite{grabisch1999axiomatic} for its simplicity and efficiency. Banzhaf Interaction is one of the most popular concepts in cooperative games~\cite{marichal2011weighted}. As shown in Fig.~\ref{ban}, it measures the additional benefits brought by the coalition compared with the costs of the lost coalitions of these players with others. When a coalition has high Banzhaf Interaction, it will also have a high contribution to the semantic similarity. Thus, we can use Banzhaf Interaction to value possible correspondence between video frames and text words for sensitive and explainable cross-modal contrast.

To this end, we propose Hierarchical Banzhaf Interaction~(HBI). Concretely, we take video frames and text words as players and the cross-modality similarity measurement as the characteristic function in the cooperative game. Then, we use the Banzhaf Interaction to represent the trend of cooperation between any set of features. Besides, to efficiently generate coalitions among game players, we propose an adaptive token merge module to cluster the original video frames (text words). By stacking token merge modules, we achieve hierarchical interaction, \ie, entity-level interactions on the frames and words, action-level interactions on the clips and phrases, and event-level interactions on the segments and paragraphs. In particular, we show that the Banzhaf Interaction index satisfies \textit{Symmetry}, \textit{Dummy}, \textit{Additivity}, and \textit{Recursivity} axiom in Sec.~\ref{analysis}. This result implies that the representation learned via Banzhaf Interaction has four properties that the features of the contrastive method do not. We find that explicitly establishing the fine-grained interpretable relationships between video and text brings a sensible improvement to already very strong video-language representation learning results. Experiment results on three text-video retrieval benchmark datasets (\textit{MSRVTT}~\cite{xu2016msr}, \textit{ActivityNet Captions}~\cite{krishna2017dense}, and \textit{DiDeMo}~\cite{anne2017localizing}) and the video question answering benchmark dataset (\textit{MSRVTT-QA}~\cite{xu2017video}) show the advantages of the proposed method. The main contributions are as follows:

\begin{figure}[tbp]
\centering
\includegraphics[width=1.\linewidth]{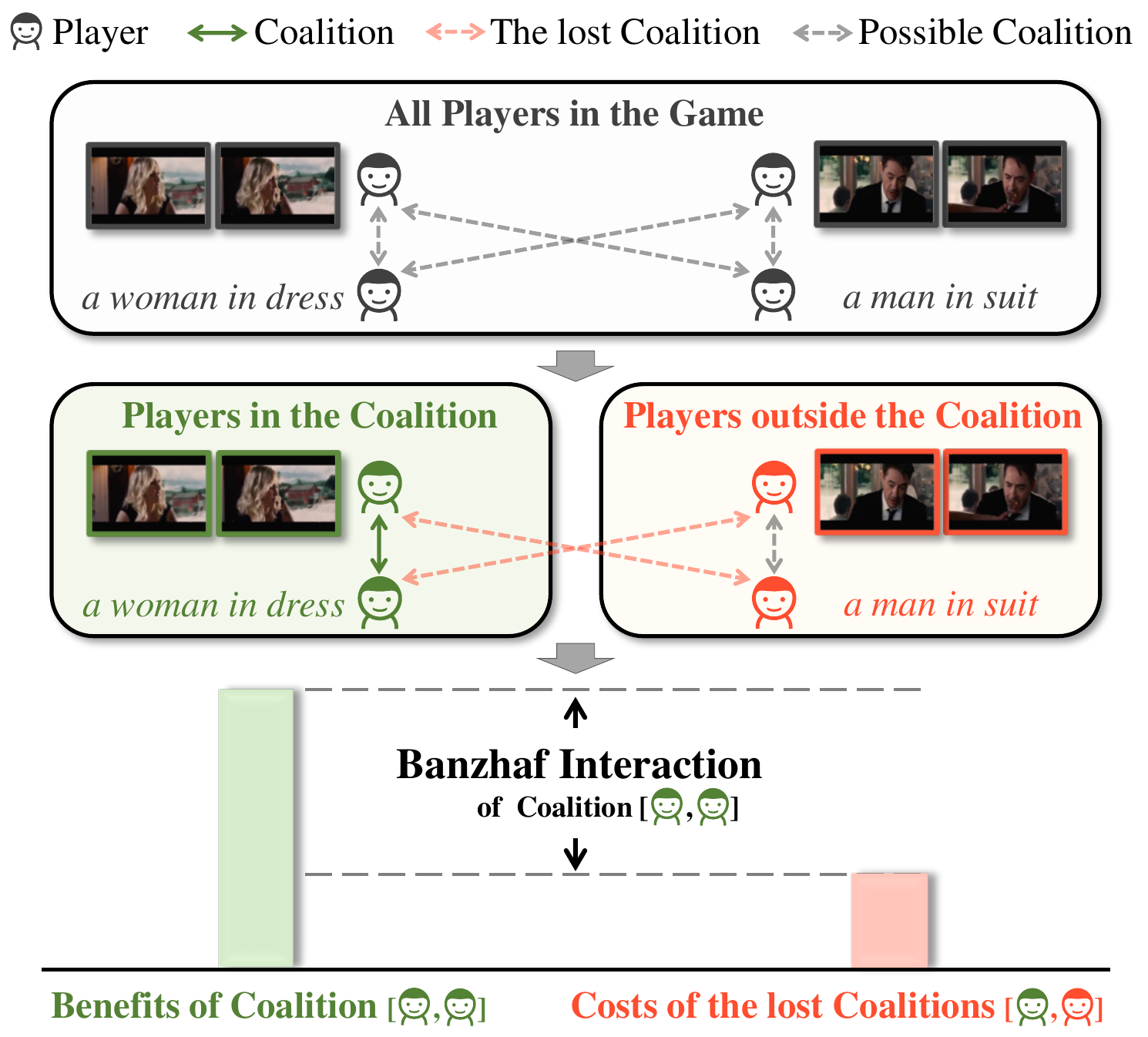}
\vspace{-1.8em}
\caption{\textbf{The intuition of Banzhaf Interaction in video-text representation learning.} We refer the reader to Eq.~\ref{BI} for the detailed formula. When some players (frames and words) form a coalition, we lose the coalitions of these players with others. In other words, the lost coalition is mutually exclusive from the target coalition. Banzhaf Interaction measures the difference between the benefits of the coalition and the costs of the lost coalitions.}
\label{ban}
\vspace{-1.em}
\end{figure}

\begin{itemize}
    \item To the best of our knowledge, we are the first to model video-language learning as a multivariate cooperative game process and propose a novel proxy training objective, which uses Banzhaf interaction to value possible correspondence between video frames and text words for sensitive and explainable cross-modal contrast.

    \item Our method achieves new state-of-the-art performance on text-video retrieval benchmarks of \textit{MSRVTT}, \textit{ActivityNet Captions} and \textit{DiDeMo}, as well as on the video-question answering task on \textit{MSRVTT-QA}.
    
    \item More encouragingly, our method can also serve as a visualization tool to promote the understanding of cross-modal interaction, which may have a far-reaching impact on the community.
\end{itemize}

\section{Related Work}
\myparagraph{Cooperative Game Theory.}
The cooperative game theory consists of a set of players with a characteristic function~\cite{osborne1994course,chalkiadakis2011computational}. The characteristic function maps each team of players to a real number which indicates the payoff obtained by all players working together to complete the task. The core of the cooperative game theory is to allocate different payoffs to game individuals fairly and reasonably. Game theory has found many applications in the field of model interpretability~\cite{datta2016algorithmic,zhang2021interpreting,yang2022explaining}, but there is little exploration in cross-modal learning. Banzhaf Interaction is one of the most popular concepts in cooperative games~\cite{marichal2011weighted}. Recently, LOUPE~\cite{li2022fine} uses two-player interaction as a vision-language pre-training task. In this paper, we design a new framework of multivariate interaction for video-text representation learning. Besides, our method can be directly co-trained with target task losses for high flexibility.

\myparagraph{Visual-Language Learning.} Recently, contrastive learning methods show great success in cross-modal tasks~\cite{jin2022expectationmaximization,li2022joint,li2022toward}, such as text-video retrieval~\cite{luo2021clip4clip,chen2020fine} and video-question answering~\cite{jin2022expectationmaximization,park2021bridge}. Text-video retrieval~\cite{wei2021universal,dong2021dual,xue2022clip} requires the model to map text and video to the same latent space, where the similarity between them can be directly calculated~\cite{chen2021learning,yang2021taco,qi2021semantics}. Video-question answering requires the model to predict an answer using visual information~\cite{zellers2021merlot,li2022joint,li2022toward}. Due to manually labeling the fine-grained relationships being unavailable, cross-modal contrastive learning cannot capture fine-grained information in a supervised manner. To this end, we model video-text as game players with multivariate cooperative game theory and propose to combine Banzhaf Interaction with cross-modal contrastive learning. In contrast to prior works, we explicitly capture the fine-grained semantic relationships between video frames and text words via Banzhaf Interaction. Then, we use these relationships as additional learning signals to improve pure contrastive learning.

\begin{figure*}[tbp]
\centering
\includegraphics[width=1.\linewidth]{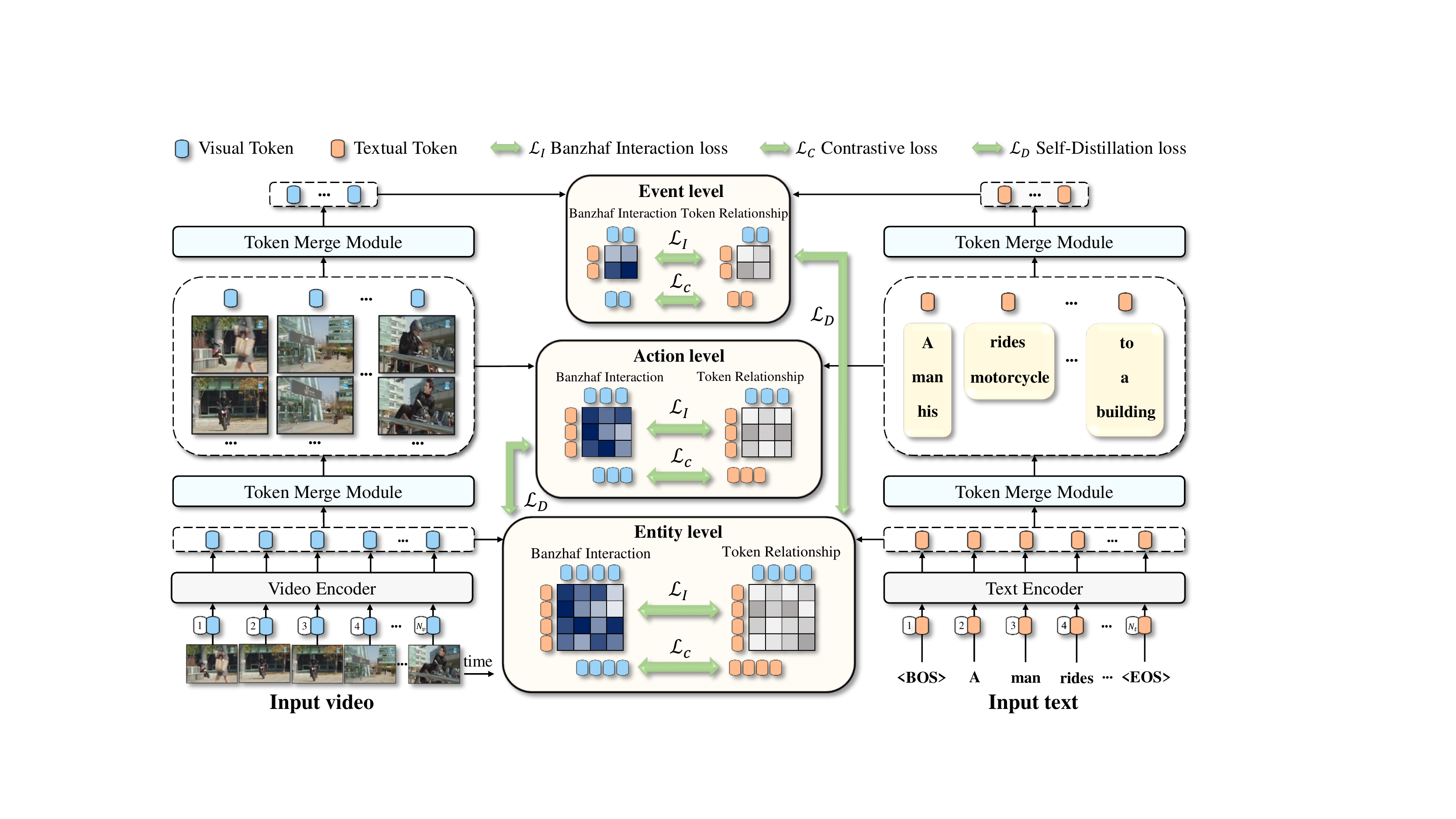}
\vspace{-1.8em}
\caption{\textbf{The overall framework of HBI.} We propose a novel proxy training objective, which uses Banzhaf interaction to value possible correspondence between video frames and text words, and enhance cross-modal representation learning. By stacking token merge modules, we achieve hierarchical interaction, \ie, entity-level interactions on the frames and words, action-level interactions on the clips and phrases, and event-level interactions on the segments and paragraphs. To improve the generalization ability, we use the additional self-distillation loss. The calculation of the exact Banzhaf Interaction is an NP-hard problem. To speed up the computation of Banzhaf Interaction for many data instances, we pre-train a tiny model to learn a mapping from a set of input features to a result (Sec.~\ref{sec:implementation}).}
\label{fig1}
\vspace{-0.5em}
\end{figure*}

\section{Method}
\subsection{Multivariate Cooperative Game Modeling}\label{3.1}
\subsubsection{Video-Language Learning}
Generally, given a corpus of video-text pairs $(\bm{v},\bm{t})$, cross-modal representation learning aims to learn a video encoder and a text encoder. The problem is formulated as a cross-modality similarity measurement $\textrm{S}_{\bm{v},\bm{t}}$ by cross-modal contrastive learning, where the matched video-text pairs are close and the mismatched pairs are away from each other.

To learn fine-grained semantic alignment, the input video $\bm{v}$ is embedded into frame sequence $\bm{V}_{f}=\{v^{i}_{f}\}^{N_v}_{i=1}$, where $N_v$ is the length of video $\bm{v}$. The input text $\bm{t}$ is embedded into word sequence $\bm{T}_{w}=\{t^{j}_{w}\}^{N_t}_{j=1}$, where $N_t$ is the length of text $\bm{t}$. Then, the alignment matrix is defined as: $A = [a_{ij}]^{N_{v}\times N_{t}}$, where $a_{ij}=\frac{(v^{i}_{f})^T t^{j}_{w}}{\Vert v^{i}_{f}\Vert \Vert t^{j}_{w}\Vert}$ represents the alignment score between the $i_{th}$ video frame and the $j_{th}$ text word. For the $i_{th}$ video frame, we calculate its maximum alignment score as $\underset{j}{\textrm{max}}\ a_{ij}$. Then, we use the weighted average maximum alignment score over all video frames as the video-to-text similarity. Similarly, we can obtain the text-to-video similarity. The total similarity score~\cite{wang2022disentangled} can be defined as: 
\begin{equation}
\textrm{S}_{\bm{v},\bm{t}}=\frac{1}{2}(\underbrace{\sum_{i=1}^{N_v} \omega_v^i\ \underset{j}{\textrm{max}}\ a_{ij}}_{\textrm{video-to-text similarity}}+\underbrace{\sum_{j=1}^{N_t} \omega_t^j\ \underset{i}{\textrm{max}}\ a_{ij}}_{\textrm{text-to-video similarity}}),
\end{equation}
where $[\omega_v^0,\omega_v^1,...,\omega_v^{N_v}]=\textrm{Softmax}(\textrm{MLP}_v(\bm{V}_{f}))$ and $[\omega_t^0,\omega_t^1,...,\omega_t^{N_t}]=\textrm{Softmax}(\textrm{MLP}_t(\bm{T}_{w}))$ are the weights of the video frames and text words, respectively. Then the cross-modal contrastive loss~\cite{van2018representation} can be formulated as:
\begin{equation}
\begin{aligned}
\mathcal{L}_{C}=-\frac{1}{2}[\frac{1}{B}\sum_{k=1}^{B}\log\frac{\exp(\textrm{S}_{\bm{v}_k,\bm{t}_k}/\tau)}{\sum_{l}^{B}\exp(\textrm{S}_{\bm{v}_k,\bm{t}_l}/\tau)} +\\  \frac{1}{B}\sum_{k=1}^{B}\log\frac{\exp(\textrm{S}_{\bm{v}_k,\bm{t}_k}/\tau)}{\sum_{l}^{B}\exp(\textrm{S}_{\bm{v}_l,\bm{t}_k}/\tau)}],
\end{aligned}
\end{equation}
where $B$ is the batch size and $\tau$ is the temperature hyper-parameter. This loss function maximizes the similarity of positive pairs and minimizes the similarity of negative pairs.

Prior works typically directly apply the cross-modal contrastive loss to optimize the similarity scores $\textrm{S}_{\bm{v},\bm{t}}$. To move a step further, we model video-text as game players with multivariate cooperative game theory to handle the uncertainty during fine-grained semantic interaction with diverse granularity, flexible combination, and vague intensity.

\subsubsection{Banzhaf Interaction}
We start by introducing notation and outlining assumptions about the cooperative game theory. Then, we review Banzhaf Interaction~\cite{grabisch1999axiomatic} for a cooperative game.

The cooperative game theory consists of a set $\mathcal{N}=\{1,2,...,n\}$ of players with a characteristic function $\phi$. The characteristic function $\phi$ maps each team of players to a real number. This number indicates the payoff obtained by all players working together to complete the task. The core of the cooperative game theory is calculating how much gain is obtained and how to distribute the total gain fairly~\cite{sun2020random}.

In a cooperative game, some players tend to form a coalition: it may happen that $\phi(\{i\})$ and $\phi(\{j\})$ are small, and at the same time $\phi(\{i,j\})$ is large. The Banzhaf Interaction~\cite{grabisch1999axiomatic} measures the additional benefits brought by the target coalition compared with the costs of the lost coalitions of these players with others. The costs of the lost coalitions can be estimated by each player in the target coalition working individually. For a coalition $\{i,j\}$, we consider $[\{i,j\}]$ as a single hypothetical player, which is the union of the players in $\{i,j\}$. Then, the reduced game is formed by removing the individual players in $\{i,j\}$ from the game and adding $[\{i,j\}]$ to the game.

\begin{myDef}
\textbf{Banzhaf Interaction~\cite{grabisch1999axiomatic}.} Given a coalition $\{i,j\} \subseteq \mathcal{N}$, the Banzhaf Interaction $\mathcal{I}([\{i,j\}])$ for the player $[\{i,j\}]$ is defined as:
\begin{equation}
\begin{aligned}
\mathcal{I}([\{i,j\}])=\!\sum_{\!\mathcal{C} \subseteq \mathcal{N} \setminus \{i,j\} }p(\mathcal{C})[\phi(\mathcal{C}\cup \{[\{i,j\}]\})+\phi(\mathcal{C})\\-\phi(\mathcal{C}\cup\{i\})-\phi(\mathcal{C}\cup\{j\})],
\end{aligned}
\label{BI}
\end{equation}
where $p(\mathcal{C})=\frac{1}{2^{n-2}}$ is the likelihood of $\mathcal{C}$ being sampled. ``$\mathcal{N} \setminus \{i,j\} $'' denotes removing $\{i,j\}$ from $\mathcal{N}$.
\end{myDef}
Intuitively, $\mathcal{I}([\{i,j\}])$ reflects the tendency of interactions inside $\{i,j\}$. The higher value of $\mathcal{I}([\{i,j\}])$ indicates that player $i$ and player $j$ cooperate closely with each other.

\subsubsection{Video-Text as Game Players}
Given features $\bm{V}_{f}=\{v^{i}_{f}\}^{N_v}_{i=1}$ and $\bm{T}_{w}=\{t^{j}_{w}\}^{N_t}_{j=1}$, fine-grained cross-modal learning aims to find semantically matched video-text feature pairs. Specifically, if a video frame and a text word have strong semantic correspondence, then they tend to cooperate with each other and contribute to the fine-grained similarity score. Thus, we can consider $\mathcal{N}=\{v^{i}_{f}\}^{N_v}_{i=1}\cup \{t^{j}_{w}\}^{N_t}_{j=1}$ as the players in the game. 

To achieve the goal of the cooperative game and cross-modal learning to be completely consistent, the characteristic function $\phi$ should meet all the following criteria: (a) the final score benefits from strongly corresponding semantic pairs $\{v^{+}_{f}, t^{+}_{w}\}$, \ie, $\phi(\mathcal{N})-\phi(\mathcal{N}\setminus\{v^{+}_{f}, t^{+}_{w}\}\cup \{[\{v^{+}_{f}, t^{+}_{w}\}]\})\textless 0$; (b) the final score is compromised by semantically irrelevant pairs $\{v^{-}_{f}, t^{-}_{w}\}$, \ie, $\phi(\mathcal{N})-\phi(\mathcal{N}\setminus\{v^{-}_{f}, t^{-}_{w}\}\cup \{[\{v^{-}_{f}, t^{-}_{w}\}]\})\textgreater 0$; (c) when there are no players to cooperate, the final score is zero, \ie, $\phi(\{v^{i}_{f}\}^{N_v}_{i=1})=\phi(\{t^{j}_{w}\}^{N_t}_{j=1})=\phi(\emptyset)=0$, where $\emptyset$ denotes the empty set.

Note that anything satisfying the above conditions can be used as the characteristic function $\phi$. For simplicity, we use cross-modality similarity measurement $\textrm{S}$ as $\phi$. Then, we can use Banzhaf Interaction to value possible correspondence between video frames and text words, and to enhance cross-modal representation learning.

\subsection{Hierarchical Banzhaf Interaction}\label{3.2}
In the following, we first introduce the simple two-player interaction between a video frame and a text word. Then, we expand the two-player interaction to the multivariate interaction via the token merge module. Fig.~\ref{fig1} illustrates the overall framework of our method.

For a coalition $\{v^{i}_{f}, t^{j}_{w}\}$, referring to Eq.~\ref{BI}, we can calculate the Banzhaf Interaction $\mathcal{I}([\{v^{i}_{f}, t^{j}_{w}\}])$. Due to the disparity in semantic similarity and interaction index, we design a prediction header to predict the fine-grained relationship $\mathcal{R}_{i,j}$ between the $i_{th}$ video frame and the $j_{th}$ text word. The prediction header consists of a convolutional layer for encoding, a self-attention module for capturing global interaction, and a convolutional layer for decoding. We provide the experiment results of the prediction header with different structures in Tab.~\ref{tab:header}.

Then, we optimize the Kullback-Leibler (KL) divergence~\cite{kullback1997information} between the $\mathcal{I}([\{v^{i}_{f}, t^{j}_{w}\}])$ and $\mathcal{R}_{i,j}$. Concretely, we define the probability distribution of the video-to-text task and the text-to-video task as:
\begin{equation}
\begin{aligned}
\mathcal{D}_{v2t}^{\mathcal{I}}&=[p_{i,1}^{\mathcal{I}},p_{i,2}^{\mathcal{I}},...,p_{i,N_t}^{\mathcal{I}}],\\
\mathcal{D}_{t2v}^{\mathcal{I}}&=[\hat p_{1,j}^{\mathcal{I}},\hat p_{2,j}^{\mathcal{I}},...,\hat p_{N_v,j}^{\mathcal{I}}],
\end{aligned}
\label{Dis}
\end{equation}
where $p_{i,j}^{\mathcal{I}}\footnotesize{=}\frac{\textrm{exp}(\mathcal{I}([\{v^{i}_{f}, t^{j}_{w}\}]))}{\sum_{k=1}^{N_t} \textrm{exp}(\mathcal{I}([\{v^{i}_{f}, t^{k}_{w}\}]))}, \hat p_{i,j}^{\mathcal{I}}\footnotesize{=}\frac{\textrm{exp}(\mathcal{I}([\{v^{i}_{f}, t^{j}_{w}\}]))}{\sum_{k=1}^{N_v} \textrm{exp}(\mathcal{I}([\{v^{k}_{f}, t^{j}_{w}\}]))}$. Similarly, the probability distribution $\mathcal{D}_{v2t}^{\mathcal{R}}$ and $\mathcal{D}_{t2v}^{\mathcal{R}}$ are calculated in the same way using $\mathcal{R}^{i,j}$, \ie, $\mathcal{D}_{v2t}^{\mathcal{R}}=[p_{i,1}^{\mathcal{R}},p_{i,2}^{\mathcal{R}},...,p_{i,N_t}^{\mathcal{R}}], \mathcal{D}_{t2v}^{\mathcal{R}}=[\hat p_{1,j}^{\mathcal{R}},\hat p_{2,j}^{\mathcal{R}},...,\hat p_{N_v,j}^{\mathcal{R}}]$, where $p_{i,j}^{\mathcal{R}}\footnotesize{=}\frac{\textrm{exp}(\mathcal{R}_{i,j})}{\sum_{k=1}^{N_t} \textrm{exp}(\mathcal{R}_{i,k})}, \hat p_{i,j}^{\mathcal{R}}\footnotesize{=}\frac{\textrm{exp}(\mathcal{R}_{i,j})}{\sum_{k=1}^{N_v} \textrm{exp}(\mathcal{R}_{k,j})}$. Finally, the Banzhaf Interaction loss $\mathcal{L}_{I}$ is defined as:
\begin{equation}
\mathcal{L}_{I}=\mathbb{E}_{\bm{v},\bm{t}} [\textrm{KL}(\mathcal{D}_{v2t}^{\mathcal{R}}\|\mathcal{D}_{v2t}^{\mathcal{I}})+\textrm{KL}(\mathcal{D}_{t2v}^{\mathcal{R}}\|\mathcal{D}_{t2v}^{\mathcal{I}})].
\end{equation}
The Banzhaf Interaction loss $\mathcal{L}_{I}$ brings the probability distributions of the output $\mathcal{R}$ of the prediction header and Banzhaf Interaction $\mathcal{I}$ close together to establish fine-grained semantic alignment between video frames and text words. In particular, it can be directly removed during inference, rendering an efficient and semantics-sensitive model.

\begin{figure}[tbp]
\centering
\includegraphics[width=1.\linewidth]{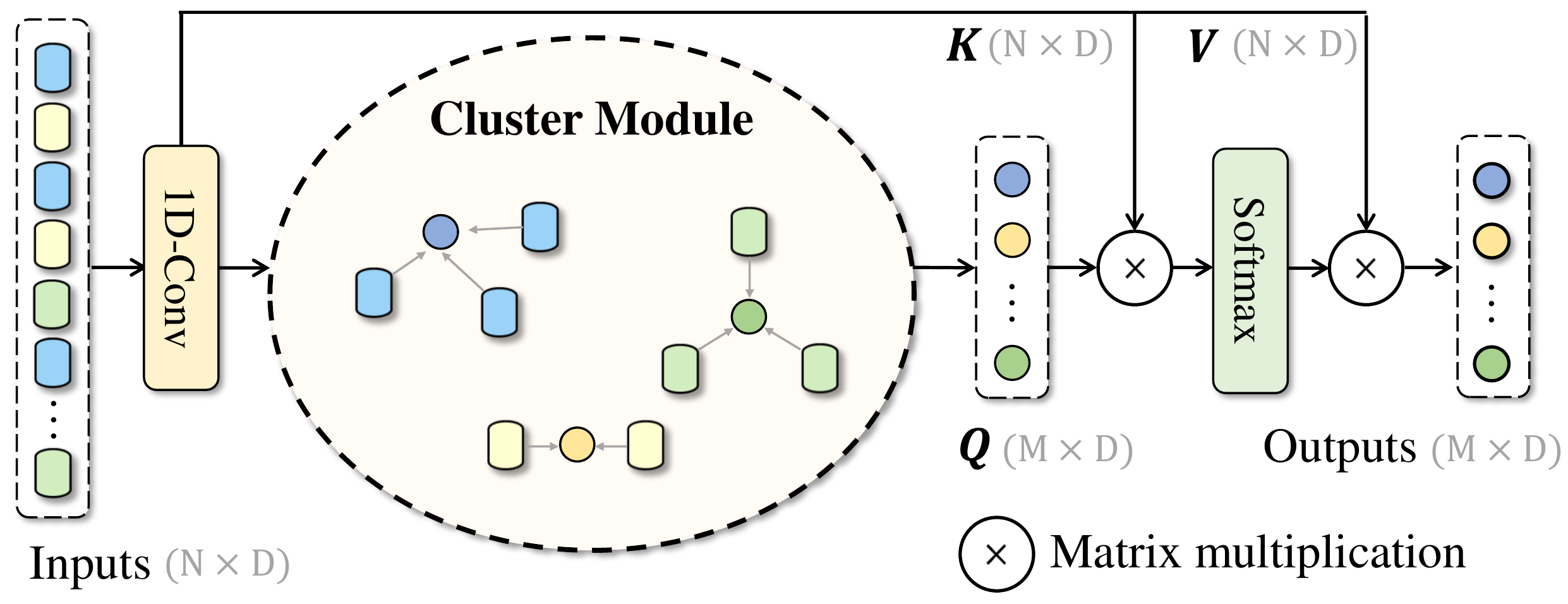}
\vspace{-1.8em}
\caption{\textbf{The token merge module.} ``1D-Conv'' denotes the one-dimensional convolutional layer. $N$ input tokens with $D$ channels are first clustered into $M$ clusters. Then, we feed the merged tokens as $Q$ and the original tokens as $K, V$ into an attention module.}
\label{fig2}
\vspace{-1.em}
\end{figure}

For multivariate interaction, an intuitive method is to compute Banzhaf Interaction on any candidate set of visual frames and text words directly. However, the number of candidate sets is too large, \ie, $2^{N_v+N_t}$. To reduce the number of candidate sets, we cluster the original visual (textual) tokens and compute the Banzhaf Interaction between the merged tokens. By stacking token merge modules, we get cross-modal interaction efficiently at different semantic levels, \ie, entity-level interactions on the frames and words, action-level interactions on the clips and phrases, and event-level interactions on the segments and paragraphs. Fig.~\ref{fig2} illustrates the framework of the token merge module.

Specifically, we utilize DPC-KNN~\cite{du2016study}, a k-nearest neighbor-based density peaks clustering algorithm, to cluster the visual (textual) tokens. Starting with the frame-level tokens $\bm{V}_{f}=\{v^{i}_{f}\}^{N_v}_{i=1}$, we first use a one-dimensional convolutional layer to enhance the temporal information between tokens. Then, we compute the local density $\rho_i$ of each token $v^{i}_{f}$ according to its $K$-nearest neighbors:
\begin{equation}
\rho_i=\textrm{exp}(-\frac{1}{K}\sum_{v^{k}_{f}\in \textrm{KNN}(v^{i}_{f})}\Vert v^{k}_{f}-v^{i}_{f} \Vert^2),
\end{equation}
where $\textrm{KNN}(v^{i}_{f})$ is the $K$-nearest neighbors of $v^{i}_{f}$. After that, we compute the distance index $\delta_i$ of each token $v^{i}_{f}$:
\begin{equation}
\delta_i=
\begin{cases}
\underset{j:\rho_j>\rho_i}{\textrm{min}} \Vert v^{k}_{f}-v^{i}_{f} \Vert^2, & \text{if $\exists j$ s.t. $\rho_j>\rho_i$.}\\
\ \ \underset{j}{\textrm{max}} \ \ \Vert v^{k}_{f}-v^{i}_{f} \Vert^2, & \text{otherwise.}
\end{cases}
\end{equation}
Intuitively, $\rho$ denotes the local density of tokens, and $\delta$ represents the distance from other high-density tokens. 

We consider those tokens with relatively high $\rho_i \times \delta_i$ as cluster centers, and then assign other tokens to the nearest cluster center according to the Euclidean distances. Inspired by~\cite{zeng2022not,rao2021dynamicvit}, we use the weighted average tokens of each cluster to represent the corresponding cluster, where the weight $W\!=\!\textrm{Softmax}(\textrm{MLP}_{w}(\bm{V}_f))$. Then, we feed the weighted average tokens as queries $Q$ and the original tokens as keys $K$ and values $V$ into an attention module. We treat the output of the attention module as features at a higher semantic level than the entity level, that is, the action-level visual tokens. Similarly, we merge the action-level tokens again to get the event-level tokens. The action-level textual tokens and event-level textual tokens are calculated in the same way.

\begin{figure*}[tbp]
\centering
\includegraphics[width=.98\linewidth]{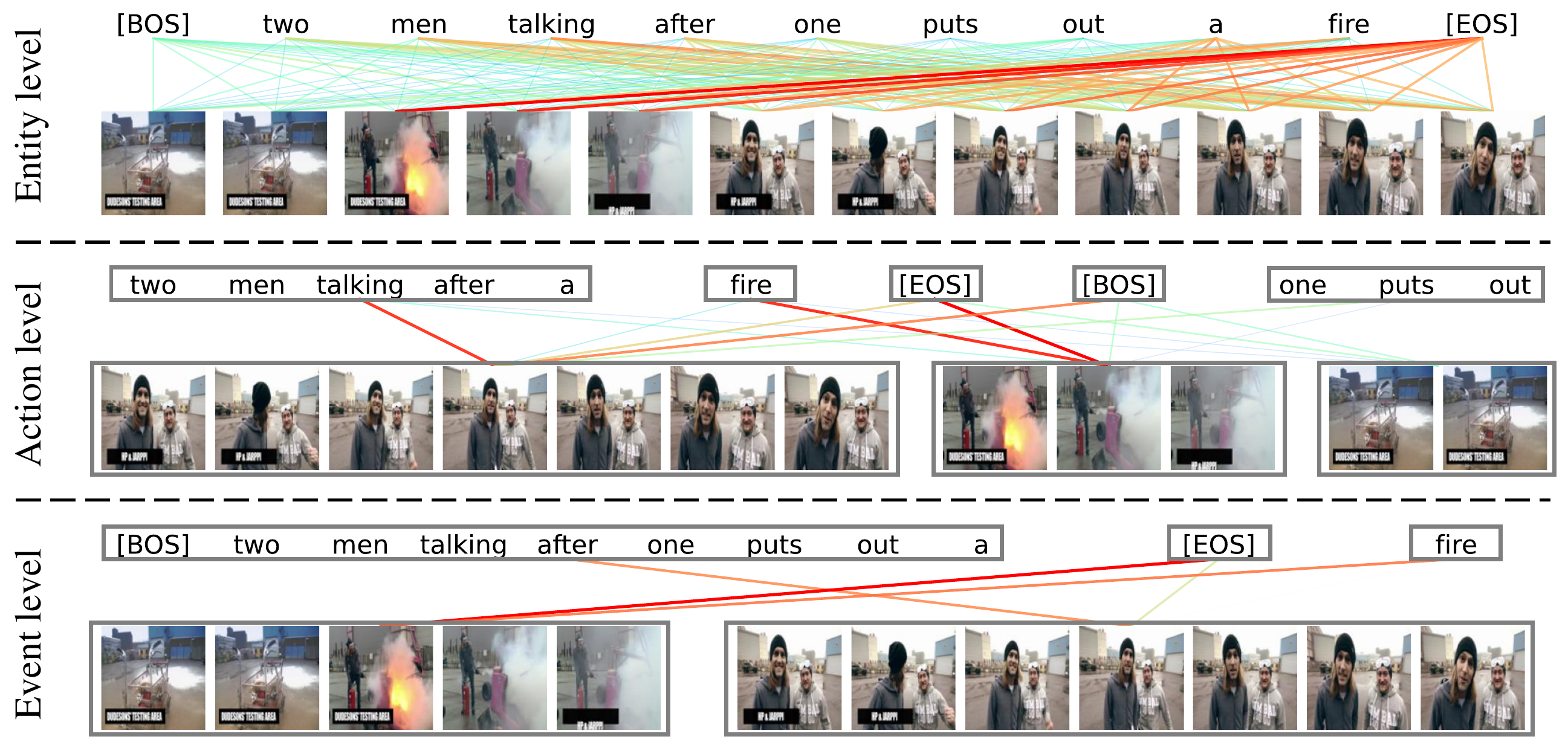}
\vspace{-0.5em}
\caption{\textbf{Visualization of the hierarchical interaction.} We take Video7060 in the MSRVTT as an example. We provide more visualizations in the Appendix. Here, the degree of confidence from high to low is represented by red, orange, green and blue lines, respectively.}
\label{fig:visualization}
\vspace{-0.8em}
\end{figure*}

\subsection{Training Objective}
Combining the cross-modal contrastive loss $\mathcal{L}_{C}$ and Banzhaf Interaction loss $\mathcal{L}_{I}$, the full objective of semantic alignment can be formulated as $\mathcal{L}=\mathcal{L}_{C}+\alpha \mathcal{L}_{I}$, where $\alpha$ is the trade-off hyper-parameter. We train the network at three semantic levels, which are shown as follows,
\begin{equation}
\mathcal{L}^e\!=\!\mathcal{L}_{C}^e\!+\!\alpha \mathcal{L}_{I}^e,\quad
\mathcal{L}^a\!=\!\mathcal{L}_{C}^a\!+\!\alpha \mathcal{L}_{I}^a,\quad
\mathcal{L}^o\!=\!\mathcal{L}_{C}^o\!+\!\alpha \mathcal{L}_{I}^o,
\label{loss0}
\end{equation}
where $\mathcal{L}^e$, $\mathcal{L}^a$, and $\mathcal{L}^o$ represent the semantic alignment loss at the entity level, action level, and event level, respectively.

To further improve the generalization ability, we optimize the additional KL divergence between the distribution among different semantic levels. We find that the entity-level similarity $\textrm{S}^e_{\bm{v},\bm{t}}$ converges first in the training process, so we distill the entity-level similarity to the other two semantic levels. The analyses and experiments are provided in Appendix.

Starting with entity-level similarity $\textrm{S}^e_{\bm{v},\bm{t}}$ distilling to action-level similarity $\textrm{S}^a_{\bm{v},\bm{t}}$, we first compute the distribution $\mathcal{D}_{v2t}^{e}$ and $\mathcal{D}_{t2v}^{e}$ by replacing $\mathcal{I}([\{v, t\}])$ with $\textrm{S}^e_{\bm{v},\bm{t}}$ in Eq.~\ref{Dis}. The distribution $\mathcal{D}_{v2t}^{a}$ and $\mathcal{D}_{t2v}^{a}$ are calculated using $\textrm{S}^a_{\bm{v},\bm{t}}$. The $\mathcal{L}_{D}^{e2a}$ loss is defined as:
\begin{equation}
\mathcal{L}_{D}^{e2a}=\mathbb{E}_{\bm{v},\bm{t}} [\textrm{KL}(\mathcal{D}_{v2t}^{a}\|\mathcal{D}_{v2t}^{e})+\textrm{KL}(\mathcal{D}_{t2v}^{a}\|\mathcal{D}_{t2v}^{e})].
\label{loss1}
\end{equation}
The $\mathcal{L}_{D}^{e2o}$ loss from entity-level similarity to event-level similarity is calculated in the same way.

The overall loss is the combination of semantically alignment losses and self-distillation losses, which is defined as:
\begin{equation}
\mathcal{L}_{total}=\underbrace{\mathcal{L}^{e}+\mathcal{L}^{a}+\mathcal{L}^{o}}_{\textrm{deep supervision}}+\beta \underbrace{(\mathcal{L}_{D}^{e2a}+\mathcal{L}_{D}^{e2o})}_{\textrm{self-distillation}},
\end{equation}
where $\beta$ is the trade-off hyper-parameter. We provide the ablation experiments for each part of the loss function in Tab.~\ref{tab:ablation}. We find that Banzhaf Interaction loss $\mathcal{L}_{I}$ significantly improves the performance, while deep supervision and self-distillation can improve the generalization ability.

\subsection{Theoretical Analysis}\label{analysis}
Similar to Banzhaf value axioms~\cite{grabisch1999axiomatic}, the following axioms convey intuitive properties that a cross-modal interaction score should satisfy.

\begin{myAxio}
Given a set $\mathcal{N}=\{1,2,...,n\}$ of players, a characteristic function $\phi:2^n \rightarrow \mathbb{R}$, and a coalition $\mathcal{C}=\{i,j\} \subseteq \mathcal{N}$, following properties are met for the interaction score $\mathcal{I}([\mathcal{C}])$.
(a) \textbf{Symmetry:} If \ \ $\forall \mathcal{S} \subseteq \mathcal{N}, \ \phi(\mathcal{S} \cup \{[\mathcal{C}]\})=\phi(\mathcal{S} \cup \{[\mathcal{C^{'}}]\}), \sum_{i \in \mathcal{C}} \phi(\mathcal{S}\cup \{i\})=\sum_{i^{'} \in \mathcal{C^{'}}} \phi(\mathcal{S}\cup \{i^{'}\})$,\ then\ $\mathcal{I}([\mathcal{C}])=\mathcal{I}([\mathcal{C^{'}}])$;
(b) \textbf{Dummy:} If \ \ $\forall \mathcal{S} \subseteq \mathcal{N},\ \phi(\mathcal{S} \cup \{[\mathcal{C}]\})=\phi(\mathcal{S}), \sum_{i \in \mathcal{C}} \phi(\mathcal{S}\cup {i})=0$,\ then\ $\mathcal{I}([\mathcal{C}])=0$;
(c) \textbf{Additivity:}  If $\phi(*)$ and $\phi^{'}(*)$ have the interaction scores $\mathcal{I}([\mathcal{C}])$ and $\mathcal{I}^{'}([\mathcal{C}])$ respectively, then the interaction score for the game with value function $\phi(*)+\phi^{'}(*)$ is $\mathcal{I}([\mathcal{C}])+\mathcal{I}^{'}([\mathcal{C}])$;
(d) \textbf{Recursivity:}  let $\mathcal{B}(*)$ denote the Banzhaf value~\cite{banzhaf1964weighted}, then $\mathcal{B}([\mathcal{C}]|\mathcal{N} \setminus \mathcal{C}\cup \{[\mathcal{C}]\})=\mathcal{B}(i|\mathcal{N} \setminus \{j\})+\mathcal{B}(j|\mathcal{N} \setminus \{i\})+\mathcal{I}([\mathcal{C}])$.
\end{myAxio}
\textbf{Symmetry} states that if changing the value of two coalitions has the same effect on the output under all values of the other variables, then both coalitions should have an identical interaction score. \textbf{Dummy} states that if changing the value of a coalition $[\mathcal{C}]$ has no effect on the output under all values of other variables, then the interaction value of $[\mathcal{C}]$ should be zero. \textbf{Additivity} states the sum of the interaction scores of the two characteristic functions is equal to the interaction score of the sum of these characteristic functions. \textbf{Recursivity} states that if the interaction is positive, then the interaction score of $[\{i,j\}]$ should be greater than simply the sum of individual values. If the interaction is negative, the interaction score of $[\{i,j\}]$ should be less than the sum.

\begin{table}[t]
\resizebox{1.\linewidth}{!}
{
\scriptsize
\begin{tabular}{l|ccc}
\toprule[1.25pt]
Method & Receptive field & Robustness & Flexibility \\ \midrule
Cosine similarity & Element level & Absolute value & Non-adjustable\\
\rowcolor{aliceblue!60} \textbf{Banzhaf Interaction} & \textbf{Set level} & \textbf{Relative value} & \textbf{Adaptable $\phi$}\\
\bottomrule[1.25pt]
\end{tabular}
}
\vspace{-.8em}
\caption{\textbf{The comparison with cosine similarity.}}
\label{advantages}
\vspace{-1.2em}
\end{table}

\newcommand{\Frst}[1]{{\textbf{#1}}}
\newcommand{\Scnd}[1]{{\underline{#1}}}
\begin{table*}[t]
\centering
\scriptsize
\subfloat[{Retrieval performance on the \textbf{Text-\textgreater{}Video} task.}]
{
\setlength{\tabcolsep}{1.3pt}
{
\begin{tabular}{lccccc}
\toprule[1.25pt]
\multicolumn{6}{c}{MSRVTT}\\
\cmidrule(rl){0-5}
Methods  & R@1$\uparrow$ & R@5$\uparrow$ & R@10$\uparrow$ & MdR$\downarrow$ & MnR$\downarrow$ \\ \midrule
MMT~\cite{gabeur2020multi} & 26.6 & 57.1 & 69.6 & 4.0 &24.0\\
T2VLAD~\cite{wang2021t2vlad} & 29.5 & 59.0 & 70.1 & 4.0 & - \\
Support-Set~\cite{patrick2021support} & 30.1 & 58.5 & 69.3 & 3.0 & - \\
CLIP4Clip~\cite{luo2021clip4clip} & 44.5 & 71.4 & 81.6 &\textbf{2.0}  & 15.3 \\
EMCL-Net~\cite{jin2022expectationmaximization} & 46.8 & 73.1 & 83.1 & \textbf{2.0} &-\\
X-Pool~\cite{gorti2022x} & 46.9 & 72.8 & 82.2 & \textbf{2.0} & 14.3 \\
TS2-Net~\cite{liu2022ts2} & 47.0 & 74.5 & \textbf{83.8} & \textbf{2.0} & 13.0 \\
\midrule
\rowcolor{aliceblue!60} \textbf{HBI (Ours)} & \textbf{48.6} & \textbf{74.6} & 83.4 & \textbf{2.0} & \textbf{12.0} \\
\bottomrule[1.25pt]
\end{tabular}
\ \quad
\begin{tabular}{lccccc}
\toprule[1.25pt]
\multicolumn{6}{c}{ActivityNet Captions}\\
\cmidrule(rl){0-5}
Methods  & R@1$\uparrow$ & R@5$\uparrow$ & R@10$\uparrow$ & MdR$\downarrow$ & MnR$\downarrow$ \\ \midrule
ClipBERT~\cite{lei2021less} & 21.3 & 49.0 & 63.5 & 6.0 & - \\
T2VLAD~\cite{wang2021t2vlad} & 23.7 & 55.5 & - & 4.0 & - \\
MMT~\cite{gabeur2020multi} & 28.7 & 61.4 & - & 3.3 & 16.0\\
Support-Set~\cite{patrick2021support} & 29.2 & 61.6 & - & 3.0 & - \\
CLIP4Clip~\cite{luo2021clip4clip} & 40.5 & 72.4 & 83.6 & \textbf{2.0}  & 7.5 \\
TS2-Net~\cite{liu2022ts2} & 41.0 & \textbf{73.6} & 84.5 & \textbf{2.0} & 8.4 \\
EMCL-Net~\cite{jin2022expectationmaximization} & 41.2 & 72.7 & - & \textbf{2.0} &-\\
\midrule
\rowcolor{aliceblue!60} \textbf{HBI (Ours)} & \Frst{42.2} & {73.0} & \Frst{84.6} & \Frst{2.0} & \Frst{6.6} \\
\bottomrule[1.25pt]
\end{tabular}
\ \quad
\begin{tabular}{lccccc}
\toprule[1.25pt]
\multicolumn{6}{c}{DiDeMo}\\
\cmidrule(rl){0-5}
Methods  & R@1$\uparrow$ & R@5$\uparrow$ & R@10$\uparrow$ & MdR$\downarrow$ & MnR$\downarrow$ \\ \midrule
FSE~\cite{zhang2018cross}  & 13.9 & 36.0 & - & 11.0 & - \\
CE~\cite{liu2019use}  & 16.1 & 41.1 & - & 8.3 & 43.7\\
ClipBERT~\cite{lei2021less} & 20.4 & 48.0 & 60.8 & 6.0 & - \\
TT-CE~\cite{croitoru2021teachtext} & 21.6 & 48.6 & 62.9 & 6.0 & - \\
Frozen~\cite{bain2021frozen} & 34.6 & 65.0 & 74.7 & 3.0 & - \\
TS2-Net~\cite{liu2022ts2} & 41.8 & 71.6 & 82.0 & \textbf{2.0} & 14.8 \\
CLIP4Clip~\cite{luo2021clip4clip} & 42.8 & 68.5 & 79.2 & \textbf{2.0}  & 18.9 \\
\midrule
\rowcolor{aliceblue!60} \textbf{HBI (Ours)} & \textbf{46.9}  & \textbf{74.9} & \textbf{82.7}  & \textbf{2.0} & \textbf{12.1}  \\
\bottomrule[1.25pt]
\end{tabular}}} 
\\ 
\vspace{0.3em}
\subfloat[{Retrieval performance on the \textbf{Video-\textgreater{}Text} task.}]
{
\setlength{\tabcolsep}{1.3pt}
\begin{tabular}{lccccc}
\toprule[1.25pt]
\multicolumn{6}{c}{MSRVTT}\\
\cmidrule(rl){0-5}
Methods  & R@1$\uparrow$ & R@5$\uparrow$ & R@10$\uparrow$ & MdR$\downarrow$ & MnR$\downarrow$ \\ \midrule
T2VLAD~\cite{wang2021t2vlad} & 31.8 & 60.0 & 71.1 & 3.0 & - \\
HiT~\cite{liu2021hit} & 32.1 & 62.7 & 74.1 & 3.0 & - \\
CLIP4Clip~\cite{luo2021clip4clip} & 42.7 & 70.9 & 80.6 & \textbf{2.0}  & 11.6 \\
X-Pool~\cite{gorti2022x} & 44.4 & 73.3 & 84.0 & \textbf{2.0} & 9.0\\
TS2-Net~\cite{liu2022ts2} & 45.3 & 74.1 & 83.7 & \textbf{2.0} & 9.2 \\
\midrule
\rowcolor{aliceblue!60} \textbf{HBI (Ours)} & \Frst{46.8} & \Frst{74.3} & \textbf{84.3} & \Frst{2.0} & \Frst{8.9} \\
\bottomrule[1.25pt]
\end{tabular}
\ \quad
\begin{tabular}{lccccc}
\toprule[1.25pt]
\multicolumn{6}{c}{ActivityNet Captions}\\
\cmidrule(rl){0-5}
Methods  & R@1$\uparrow$ & R@5$\uparrow$ & R@10$\uparrow$ & MdR$\downarrow$ & MnR$\downarrow$ \\ \midrule
HSE~\cite{zhang2018cross} & 18.7 & 48.1 & - & - & - \\
T2VLAD~\cite{wang2021t2vlad} & 24.1 & 56.6 & - & 4.0 & - \\
Support-Set~\cite{patrick2021support} & 28.7 & 60.8 & - & \textbf{2.0} & - \\
MMT~\cite{gabeur2020multi} & 28.9 & 61.1 & - & 4.0 & 17.1\\
CLIP4Clip~\cite{luo2021clip4clip} & 41.4 & \Frst{73.7} & 85.3 & \textbf{2.0}  & 6.7 \\
\midrule
\rowcolor{aliceblue!60} \textbf{HBI (Ours)} & \Frst{42.4}  & 73.0 & \textbf{86.0} & \Frst{2.0} & \Frst{6.5} \\
\bottomrule[1.25pt]
\end{tabular}
\quad \
\begin{tabular}{lccccc}
\toprule[1.25pt]
\multicolumn{6}{c}{DiDeMo}\\
\cmidrule(rl){0-5}
Methods  & R@1$\uparrow$ & R@5$\uparrow$ & R@10$\uparrow$ & MdR$\downarrow$ & MnR$\downarrow$ \\ \midrule
FSE~\cite{zhang2018cross} & 13.1 & 33.9 & - & 12.0 & -\\
S2VT~\cite{venugopalan2014translating} & 13.2 & 33.6 & - & 15.0 & - \\
CE~\cite{liu2019use}  & 15.6 & 40.9 & - & 8.2 & 42.4 \\
TT-CE~\cite{croitoru2021teachtext} & 21.1 & 47.3 & 61.1 & 6.3 & -\\
CLIP4Clip~\cite{luo2021clip4clip} & 41.4 & 68.2 & 79.1 & \textbf{2.0}  & 12.4 \\
\midrule
\rowcolor{aliceblue!60} \textbf{HBI (Ours)} & \textbf{46.2}  & \textbf{73.0} & \textbf{82.7} & \textbf{2.0} & \textbf{8.7} \\
\bottomrule[1.25pt]
\end{tabular}
}
\vspace{-1.em}
\caption{\textbf{Comparisons to current state-of-the-art methods on the MSRVTT \cite{xu2016msr}, ActivityNet Captions~\cite{krishna2017dense} and DiDeMo~\cite{anne2017localizing} datasets.} ``$\uparrow$'' denotes that higher is better. ``$\downarrow$'' denotes that lower is better. All results in this table do not use inverted softmax~\cite{bogolin2022cross}.}
\label{Comparisons to State-of-the-arts}
\vspace{-1.em}
\end{table*}

\begin{myTheo}
The Banzhaf Interaction index satisfies \textbf{Symmetry}, \textbf{Dummy}, \textbf{Additivity} and \textbf{Recursivity} axiom.
\label{Theorem 1}
\end{myTheo}
We refer the reader to Appendix for more detail about Theorem~\ref{Theorem 1}. This result implies that the representation learned via Banzhaf Interaction has four properties that the features of the contrastive method do not. Besides, we compare Banzhaf Interaction and cosine similarity in Tab.~\ref{advantages}, mainly in three aspects. \textbf{(1) {Global receptive field.}} In contrast to cosine similarity, which only operates at the element level, Banzhaf Interaction operates at the set level to leverage the global context. \textbf{(2) {Robustness.}} Cosine similarity fluctuates by visual and language style. In contrast, Banzhaf Interaction measures the relative value of benefit and opportunity cost to be robust to the style deviation. {\textbf{(3) Flexibility.}} Our framework can use other characteristic functions $\phi$ besides similarity, which is left for future work to explore. Therefore, Banzhaf Interaction is a promising interaction score to enhance cross-modal representation learning.

\section{Experiments}
\subsection{Experimental Settings}
\myparagraph{Datasets.} \textbf{MSRVTT}~\cite{xu2016msr} contains 10K YouTube videos, each with 20 text descriptions. We follow the training protocol in \cite{liu2019use,gabeur2020multi} and evaluate on the 1K-A testing split~\cite{yu2018a}. \textbf{ActivityNet Captions}~\cite{krishna2017dense} consists of densely annotated temporal segments of 20K YouTube videos. We use the 10K training split to train the model and report the performance on the 5K ``val1'' split.  \textbf{DiDeMo}~\cite{anne2017localizing} contains 10K videos annotated 40K text descriptions. We follow the training and evaluation protocol in \cite{luo2021clip4clip}. \textbf{MSRVTT-QA}~\cite{xu2017video} is based on the MSRVTT and has 243K VideoQA pairs.

\myparagraph{Metrics.} We choose Recall at rank K (R@K), Median Rank (MdR), and mean rank (MnR)~\cite{gabeur2020multi} to evaluate the retrieval performance. We choose answer accuracy to evaluate the video question answering performance.

\myparagraph{Implementation Details.}\label{sec:implementation}
Since the calculation of the exact Banzhaf Interaction is an NP-hard problem~\cite{matsui2001np}, existing methods mainly use sampling-based methods~\cite{leech2002computation,bachrach2010approximating} to obtain unbiased estimates. To speed up the computation of Banzhaf Interaction for many data instances, we pre-train a tiny model to learn a mapping from a set of input features to a result using MSE loss. The tiny model consists of 2 CNN layers and a self-attention layer. The input is the similarity matrix of video frames and text tokens, and the output is the estimation of Banzhaf Interaction. We refer the reader to Appendix for the details. For text-video retrieval, we utilize the CLIP (ViT-B/32)~\cite{radford2021learning} as the pre-trained model. For video question answering, we use the target vocabulary and train a fully connected layer on top of the final language features to classify the answer. More details are in the Appendix.

\subsection{Comparison with State-of-the-art}
In Tab.~\ref{Comparisons to State-of-the-arts}, we show the results of our method on MSRVTT, ActivityNet Captions, and DiDeMo datasets. Our model consistently outperforms the recently proposed state-of-the-art methods on both text-to-video retrieval and video-to-text retrieval tasks. Tab.~\ref{videoqa} shows the results of our method for video-question answering. Massive experiments on text-video retrieval and video-question answering tasks demonstrate the superiority and flexibility of our method.

\newfloat{figtab}{htb}{fgtb}
\makeatletter
  \newcommand\figcaption{\def\@captype{figure}\caption}
  \newcommand\tabcaption{\def\@captype{table}\caption}
\makeatother

\begin{figure*}[t]
\begin{minipage}[b]{0.21\textwidth}
\centering
\scriptsize
\centering
\setlength{\tabcolsep}{.pt}
\begin{tabular}{lc}
\toprule[1.25pt]
{Methods} & Accuracy (\%)$\uparrow$\\ 
 \midrule
VQA-T~\cite{yang2021just}  & 41.5 \\
SiaSamRea~\cite{yu2021learning} & 41.6 \\
MERLOT~\cite{zellers2021merlot} & 43.1 \\
Co-Tokenization~\cite{piergiovanni2022video} & 45.7 \\
EMCL-QA~\cite{jin2022expectationmaximization} & 45.8\\
\midrule
\rowcolor{aliceblue!60} \textbf{HBI (Ours)} & \textbf{46.2} \\
\bottomrule[1.25pt]
\end{tabular}
\vspace{-1.3em}
\tabcaption{\textbf{Video-question answering performance on MSRVTT-QA dataset.}}
\label{videoqa}
\end{minipage}
\hfill
\begin{minipage}[b]{0.25\textwidth}
\centering
{
\centering
\scriptsize
\setlength{\tabcolsep}{0.8pt}
\begin{tabular}{lcccc}
\toprule[1.25pt]
\multirow{2}{*}{Method} &  \multicolumn{4}{c}{{Text-\textgreater{}Video}}  \\
\cline{2-5}
  & R@1$\uparrow$ & R@5$\uparrow$ & R@10$\uparrow$ & MnR$\downarrow$ \\ \midrule
Baseline  & 46.6 & 73.1 & 83.0 & 13.3 \\
\midrule
MLP  & 47.2 & 73.7 & 83.5 & 12.3\\
CNN  & 47.3 & 73.5 & \textbf{83.7} & 12.2\\
MLP+SA  & 46.6 & 74.0 & \textbf{83.7} & 12.3\\
\rowcolor{aliceblue!60} CNN+SA   & \textbf{48.6} & \textbf{74.6} & 83.4 & \textbf{12.0} \\ 
\bottomrule[1.25pt]
\end{tabular}
}
\vspace{-0.8em}
\tabcaption{\textbf{Effect of the prediction header on MSRVTT dataset.} ``SA'' is the self-attention module.}
\label{tab:header}
\end{minipage}
\hfill
  \begin{minipage}[b]{0.50\textwidth}
    \centering
    \vspace{-0.5em}
    \includegraphics[width=1.\textwidth]{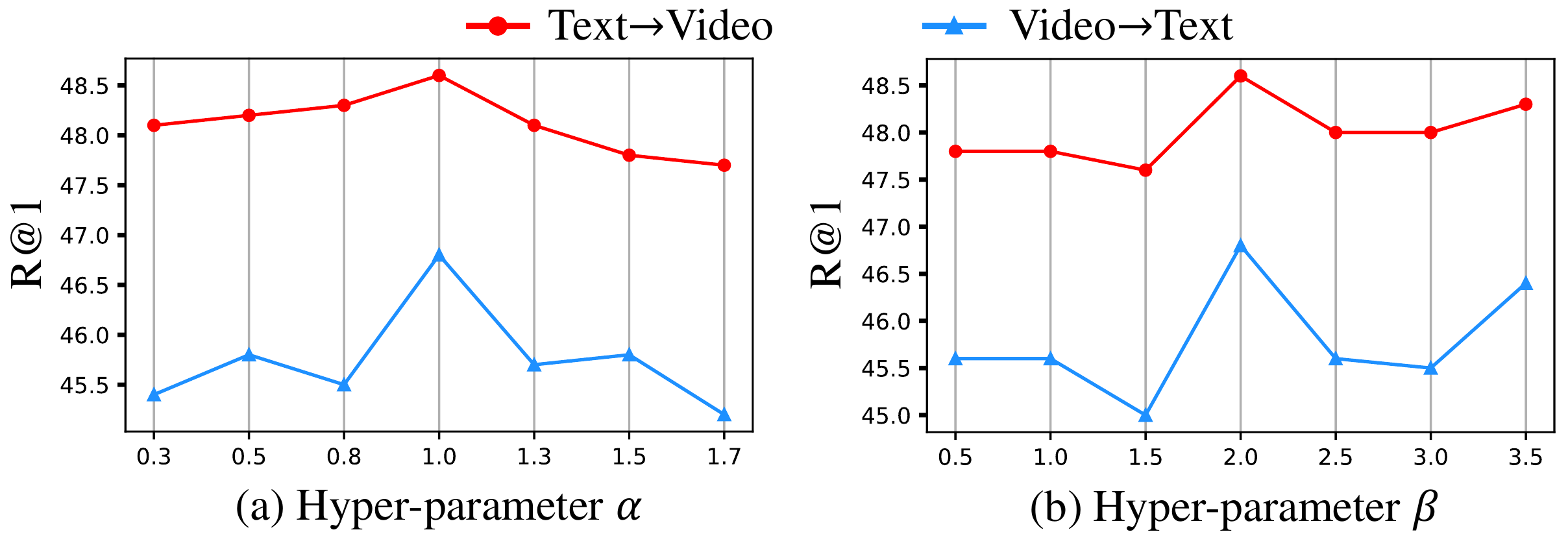}
    \vspace{-1.8em}
    \caption{\textbf{Effect of the hyper-parameters on MSRVTT dataset.} $\alpha$ and $\beta$ are the hyper-parameters in Eq.~\ref{loss0} and Eq.~\ref{loss1}, respectively.}
\label{fig:Hyper-parameter}
 \end{minipage}
\vspace{-.5em}
\end{figure*}

\begin{table*}[t]
\scriptsize
\begin{minipage}[b]{0.4\textwidth}
\centering
{
\scriptsize
\setlength{\tabcolsep}{1.5pt}
\begin{tabular}{ccccccc}
\toprule[1.25pt]
{$\mathcal{L}_{I}$ Banzhaf} & {Deep} & {$\mathcal{L}_{D}$ Self} & \multicolumn{4}{c}{{Text-\textgreater{}Video}}  \\  \cline{4-7}
{Interaction} & {Supervision} & {Distillation} & R@1$\uparrow$ & R@5$\uparrow$ & R@10$\uparrow$ & MnR$\downarrow$\\ \midrule
& &  & 46.6 & 73.1 & 83.0 & 13.3\\ \midrule
 \tiny{\Checkmark} & &  & 47.4 & 74.2 & 82.8 & 12.1\\\
 & \tiny{\Checkmark} &  & 47.2 & 74.1 & 82.6 & 12.0\\
 & \tiny{\Checkmark} & \tiny{\Checkmark}  & 47.6 & 73.8 & 83.2 & \textbf{11.9}\\
\tiny{\Checkmark}  & \tiny{\Checkmark} & & 48.2 & 73.0 & 83.1 & 12.0\\
\rowcolor{aliceblue!60} \tiny{\Checkmark} & \tiny{\Checkmark} & \tiny{\Checkmark}  & \textbf{48.6} & \textbf{74.6} & \textbf{83.4} & 12.0\\
\bottomrule[1.25pt]
\end{tabular}
\vspace{-.8em}
\caption{\textbf{Ablation study about the importance of each part of our method on MSRVTT dataset.}}
\label{tab:ablation}
}
\end{minipage}
\hfill
\begin{minipage}[b]{0.34\textwidth}
\centering
{
\centering
\scriptsize
\setlength{\tabcolsep}{1.3pt}
\begin{tabular}{ccccccccc}
\toprule[1.25pt]
\multirow{2}{*}{$N^a_v$} & \multirow{2}{*}{$N^o_v$} & \multirow{2}{*}{$N^a_t$} & \multirow{2}{*}{$N^o_t$} &  \multicolumn{5}{c}{{Text-\textgreater{}Video}}  \\
\cline{5-9}
& & & & R@1$\uparrow$ & R@5$\uparrow$ & R@10$\uparrow$ & Rsum$\uparrow$ & MnR$\downarrow$ \\ \midrule
- & - & - & -  & {47.5} & {73.7} & 83.0 & 204.2 & \textbf{12.0}\\
\midrule
9 & 3 & 18 & 4 & 48.2 & \textbf{75.2} & 82.7 & 206.1 & 12.4 \\
6 & 3 & 12 & 4 & 48.3 & 74.3 & 83.1 & 205.7 & 12.3 \\
6 & 2 & 12 & 3 & \textbf{48.7} & 74.5 & 82.6 & 205.8 & 12.2 \\
\rowcolor{aliceblue!60} 3 & 2 & 6 & 3 & 48.6 & 74.6 & \textbf{83.4} & \textbf{206.6} & \textbf{12.0} 
\\ \bottomrule[1.25pt]
\end{tabular}
}
\vspace{-1.em}
\caption{\textbf{The efficiency of the cluster module.} {$N_{-}^{a}$} and {$N_{-}^{o}$} denote the number of clusters at the action level and event level, respectively.}
\label{cluster}
\end{minipage}
\hfill
\begin{minipage}[b]{0.22\textwidth}
\scriptsize
\centering
\setlength{\tabcolsep}{1.2pt}
\begin{tabular}{lcc}
\toprule[1.25pt]
\multirow{2}{*}{Method} & Iteration & Inference \\ 
& Time$\downarrow$ & Time$\downarrow$ \\
 \midrule
CLIP4Clip~\cite{luo2021clip4clip} & \textbf{1.63} s & \textbf{16.28} s\\
DRL~\cite{wang2022disentangled} & 1.65 s & 16.74 s\\
EMCL-Net~\cite{jin2022expectationmaximization} & 1.72 s & 17.68 s\\
TS2-Net~\cite{liu2022ts2} & 2.57 s & 19.91 s\\
 \midrule
Baseline & 2.06 s & 18.06 s\\ 
HBI (Ours)  & 3.14 s  & 19.17 s\\
\bottomrule[1.25pt]
\end{tabular}
\vspace{-.5em}
\caption{\textbf{Time consumption on MSRVTT dataset.}}
\label{tab:time}
\end{minipage}
\vspace{-1.0em}
\end{table*}

\subsection{Ablation Study}
\myparagraph{Effect of the prediction header of $\mathcal{R}$.}\label{prediction header} 
To explore the impact of the structure of the prediction header on our method, we compare several popular structures in Tab.~\ref{tab:header}. We find that the combination of CNN and attention (``CNN+SA'') can capture both local and global interaction, so it is beneficial for predicting the fine-grained relationship.

\myparagraph{Ablation about components.} As shown in Tab.~\ref{tab:ablation}, Banzhaf Interaction boosts the baseline with the improvement up to 0.8\% at R@1. Moreover, deep supervision and self-distillation significantly improve the generalization ability. Our full model achieves the best performance and outperforms the baseline by 2.0\% at R@1 for text-to-video retrieval. This demonstrates that the three parts are beneficial for aligning videos and texts.

\myparagraph{The efficiency of the cluster module.} 
The ablation results are provided in Tab.~\ref{cluster}. $N_{v}^{-}$ and $N_{t}^{-}$ denote the number of visual and textual clusters, respectively. The first row represents the baseline without the cluster module. We find that large numbers of clusters may make similar tokens classified in different clusters. From Tab.~\ref{cluster}, we take the $\{N^a_v, N^o_v, N^a_t, N^o_t\}$ as $\{3, 2, 6, 3\}$ to get the best performance on the sum of recall at rank $\{1,5,10\}$ (Rsum).

\myparagraph{The efficiency of our method.}
In Tab.~\ref{tab:time}, we calculate iteration time and inference time using two Tesla V100 GPUs on MSRVTT dataset. Since the Banzhaf Interaction can be removed during inference, our method only takes additional 1s for processing the test set. This result demonstrates the superiority of our efficient design.

\myparagraph{Parameter sensitivity.}
The parameter $\alpha$ is the hyper-parameter that trades off $\mathcal{L}_{C}$ and $\mathcal{L}_{I}$. We evaluate the scale range setting $\alpha \in [0.3, 1.7]$ as shown in Fig.~\ref{fig:Hyper-parameter}a. From Fig.~\ref{fig:Hyper-parameter}a, we adopt $\alpha = 1.0$ to achieve the best performance. In Fig.~\ref{fig:Hyper-parameter}b, we show the influence of the hyper-parameter $\beta$. We evaluate the scale range setting $\beta \in [0.5, 3.5]$. We find that the model achieves the best performance at $\beta=2.0$, so we set $\beta=2.0$ as default in practice.

\subsection{Qualitative Analysis}
To better understand the proposed method, we show the visualization of the hierarchical interaction in Fig.~\ref{fig:visualization}. We find that the semantic similarities between coalitions are generally higher than the semantic similarities between individual frames and individual words. For example, the coalition ``\{two, men, talking, after, a\}'' has a high semantic similarity with the video coalition representing the men talking action. On the contrary, when these words interact with the corresponding frame as individuals, they show low semantic similarity. Interestingly, the model uses the word ``fire'' instead of the phrase ``one puts out a fire'' to understand the video-text pair. This is due to insufficient training data, the model can not understand the low-frequency phrase. The visualization illustrates that the proposed method can be used as a tool for visualizing the cross-modal interaction and help us understand the cross-modal model.

\section{Conclusion}
In this paper, we creatively model cross-modal representation learning as a multivariate cooperative game by formulating video and text as players in a cooperative game. Specifically, we propose Hierarchical Banzhaf Interaction~(HBI) to value possible correspondence between video frames and text words for sensitive and explainable cross-modal contrast. Although manually labeling the fine-grained relationships between videos and text is unavailable, our method shows a promising alternative to obtaining fine-grained labels based on Banzhaf Interaction. More encouragingly, our method can also serve as a visualization tool to promote the understanding of cross-modal interaction.

\noindent \textbf{Acknowledgements.} This work was supported in part by the National Key R\&D Program of China (No. 2022ZD0118201), Natural Science Foundation of China (No. 61972217, 32071459, 62176249, 62006133, 62271465), and the Natural Science Foundation of Guangdong Province in China (No. 2019B1515120049).

{\small
\bibliographystyle{ieee_fullname}
\bibliography{bib}
}

\appendix
\renewcommand{\thetable}{\Alph{table}}
\renewcommand{\theequation}{\Alph{equation}}
\renewcommand{\thefigure}{\Alph{figure}}
\setcounter{table}{0}
\setcounter{section}{0}
\setcounter{figure}{0}
\setcounter{equation}{0}
\setcounter{myDef}{0}
\setcounter{myAxio}{0}
\setcounter{myTheo}{0}

\section{Datasets and Implementation Details}
\subsection{Datasets}
\myparagraph{MSRVTT.} MSRVTT~\cite{xu2016msr} contains 10K YouTube videos, each with 20 text descriptions. We follow the training protocol in \cite{liu2019use,gabeur2020multi} and evaluate on text-to-video and video-to-text search tasks on the 1K-A testing split with 1K video or text candidates defined by \cite{yu2018a}.

\myparagraph{ActivityNet Captions.} ActivityNet Captions~\cite{krishna2017dense} consists densely annotated temporal segments of 20K YouTube videos. Following~\cite{gabeur2020multi,patrick2021support,wang2021t2vlad}, we concatenate descriptions of segments in a video to construct ``video-paragraph'' for retrieval. We use the 10K training split to finetune the model and report the performance on the 5K ``val1'' split.

\myparagraph{DiDeMo.} DiDeMo~\cite{anne2017localizing} contains 10K videos annotated 40K text descriptions. We concatenate descriptions of segments in a video to construct ``video-paragraph'' for retrieval. We follow the training and evaluation protocol in~\cite{luo2021clip4clip}.

\myparagraph{MSRVTT-QA.} MSRVTT-QA~\cite{xu2017video} is based on the MSRVTT dataset and has 243K VideoQA pairs.

\subsection{Implementation Details}
For fair comparisons, we follow common practice~\cite{luo2021clip4clip,cheng2021improving,wang2022disentangled} to extract the video representations of input videos and the language representations of input texts. In detail, for video representations, we first extract the frames from the video clip as the input sequence of video. Then we use ViT~\cite{dosovitskiy2021an} to encode the frame sequence, by exploiting the transformer~\cite{vaswani2017attention,li2022locality} architecture to model the interactions between image patches. Followed by the CLIP~\cite{radford2021learning}, the output from the [class] token is used as the frame embedding. Finally, we obtain the video representation $\bm{V}_{f}=\{v^{i}_{f}\}^{N_v}_{i=1}$. For text representation, we directly use the text encoder of CLIP to acquire the text representation $\bm{T}_{w}=\{t^{j}_{w}\}^{N_t}_{j=1}$.

The dimension of the feature is 512. The temporal transformer is composed of 4-layer blocks, each including 8 heads and 512 hidden channels. The temporal position embedding and parameters are initialized from the CLIP’s text encoder. We use the Adam optimizer~\cite{kingma2014adam} and set the temperature $\tau$ to 0.01. The initial learning rate is 1e-7 for text encoder and video encoder and 1e-3 for other modules.

For text-video retrieval, we utilize the CLIP (ViT-B/32)~\cite{radford2021learning} as the pre-trained model. The frame length and caption length are 12 and 24 for MSRVTT. The network is optimized with the batch size of 128 in 5 epochs. We set the caption length to 64 for ActivityNet Captions and DiDeMo. 

For video question answering~\cite{jin2022expectationmaximization,park2021bridge,ye2023fits}, we use the target vocabulary and train a fully connected layer on top of the final language features to classify the answer. The frame length and question length are 12 and 32 for MSRVTT-QA. The network is optimized with the batch size of 32 in 5 epochs.

\section{Proof of Theorem 1}
We start by reviewing Banzhaf Values~\cite{banzhaf1964weighted} and Banzhaf Interaction~\cite{grabisch1999axiomatic} for a cooperative game.

The cooperative game theory consists of a set $\mathcal{N}=\{1,2,...,n\}$ of players with a characteristic function $\phi:2^n \rightarrow \mathbb{R}$. The characteristic function $\phi$ maps each team of players to a real number. This number indicates the payoff obtained by all players working together to complete the task. The core of the cooperative game theory is calculating how much gain is obtained and how to distribute the total gain fairly~\cite{sun2020random}.

\myparagraph{Banzhaf Values.} The Banzhaf value~\cite{banzhaf1964weighted} is one of the most important solution concepts in cooperative games. Formally, the Banzhaf value measures the average marginal contribution of each player across all permutations of the players. It is the unbiased estimation of the importance or contribution of each player in a cooperative game~\cite{dubey1975uniqueness}, and has thus found many applications from estimating feature importance to pruning neural networks. Given a set $\mathcal{N}=\{1,2,...,n\}$ of players and a characteristic function $\phi:2^n \rightarrow \mathbb{R}$, the Banzhaf value $\mathcal{B}(i|\mathcal{N})$ for player $i$ is defined as the average marginal contribution of player $i$ to all possible coalitions $\mathcal{C} \subseteq \mathcal{N}$ that are formed without $i$:
\begin{equation}
\mathcal{B}(i|\mathcal{N})=\sum_{\mathcal{C} \subseteq \mathcal{N} \setminus \{i\}}p(\mathcal{C})(\phi(\mathcal{C}\cup \{i\})-\phi(\mathcal{C})),
\end{equation}
where $p(\mathcal{C})=\frac{1}{2^{n-1}}$ is the likelihood of $\mathcal{C}$ being sampled. ``$\mathcal{N} \setminus \{i\} $'' denotes removing $\{i\}$ from $\mathcal{N}$.

\myparagraph{Banzhaf Interaction.} In a cooperative game, some players tend to form a coalition: it may happen that $\phi(\{i\})$ and $\phi(\{j\})$ are small and at the same time $\phi(\{i,j\})$ is large. The Banzhaf Interaction~\cite{grabisch1999axiomatic} measures the additional benefits brought by the coalition compared with the costs of the lost interactions of these players with others. For a coalition $\{i,j\}$, we consider $[\{i,j\}]$ as a single hypothetical player, which is the union of the players in $\{i,j\}$. Then, the reduced game is formed by removing the individual players in $\{i,j\}$ from the game and adding $[\{i,j\}]$ to the game. 

\begin{myDef}
\textbf{Banzhaf Interaction~\cite{grabisch1999axiomatic}.} Given a coalition $\{i,j\} \subseteq \mathcal{N}$, the Banzhaf Interaction $\mathcal{I}([\{i,j\}])$ for the player $[\{i,j\}]$ is defined as:
\begin{equation}
\begin{aligned}
\mathcal{I}([\{i,j\}])\!=\!\sum_{\!\mathcal{C} \subseteq \mathcal{N} \setminus \{i,j\} }p(\mathcal{C})[\phi(\mathcal{C}\cup \{[\{i,j\}]\})+\phi(\mathcal{C})\\-\phi(\mathcal{C}\cup\{i\})-\phi(\mathcal{C}\cup\{j\})],
\end{aligned}
\label{apendix:BI}
\end{equation}
where $p(\mathcal{C})=\frac{1}{2^{n-2}}$ is the likelihood of $\mathcal{C}$ being sampled. ``$\mathcal{N} \setminus \{i,j\} $'' denotes removing $\{i,j\}$ from $\mathcal{N}$.
\end{myDef}

Similar to Banzhaf value axioms~\cite{grabisch1999axiomatic}, the following axioms convey intuitive properties that a cross-modal interaction score should satisfy.

\begin{myAxio}
Given a set $\mathcal{N}=\{1,2,...,n\}$ of players, a characteristic function $\phi:2^n \rightarrow \mathbb{R}$, and a coalition $\mathcal{C}=\{i,j\} \subseteq \mathcal{N}$, following properties are met for the interaction score $\mathcal{I}([\mathcal{C}])$.
(a) \textbf{Symmetry:} If \ \ $\forall \mathcal{S} \subseteq \mathcal{N}, \ \phi(\mathcal{S} \cup \{[\mathcal{C}]\})=\phi(\mathcal{S} \cup \{[\mathcal{C^{'}}]\}), \sum_{i \in \mathcal{C}} \phi(\mathcal{S}\cup \{i\})=\sum_{i^{'} \in \mathcal{C^{'}}} \phi(\mathcal{S}\cup \{i^{'}\})$,\ then\ $\mathcal{I}([\mathcal{C}])=\mathcal{I}([\mathcal{C^{'}}])$;
(b) \textbf{Dummy:} If \ \ $\forall \mathcal{S} \subseteq \mathcal{N},\ \phi(\mathcal{S} \cup \{[\mathcal{C}]\})=\phi(\mathcal{S}), \sum_{i \in \mathcal{C}} \phi(\mathcal{S}\cup {i})=0$,\ then\ $\mathcal{I}([\mathcal{C}])=0$;
(c) \textbf{Additivity:}  If $\phi(*)$ and $\phi^{'}(*)$ have the interaction scores $\mathcal{I}([\mathcal{C}])$ and $\mathcal{I}^{'}([\mathcal{C}])$ respectively, then the interaction score for the game with value function $\phi(*)+\phi^{'}(*)$ is $\mathcal{I}([\mathcal{C}])+\mathcal{I}^{'}([\mathcal{C}])$;
(d) \textbf{Recursivity:}  let $\mathcal{B}(*)$ denote the Banzhaf value, then $\mathcal{B}([\mathcal{C}]|\mathcal{N} \setminus \mathcal{C}\cup \{[\mathcal{C}]\})=\mathcal{B}(i|\mathcal{N} \setminus \{j\})+\mathcal{B}(j|\mathcal{N} \setminus \{i\})+\mathcal{I}([\mathcal{C}])$.
\end{myAxio}

\begin{myTheo}
The Banzhaf Interaction index satisfies \textbf{Symmetry}, \textbf{Dummy}, \textbf{Additivity} and \textbf{Recursivity} axiom.
\end{myTheo}

\begin{figure*}[tbp]
\centering
\includegraphics[width=.98\linewidth]{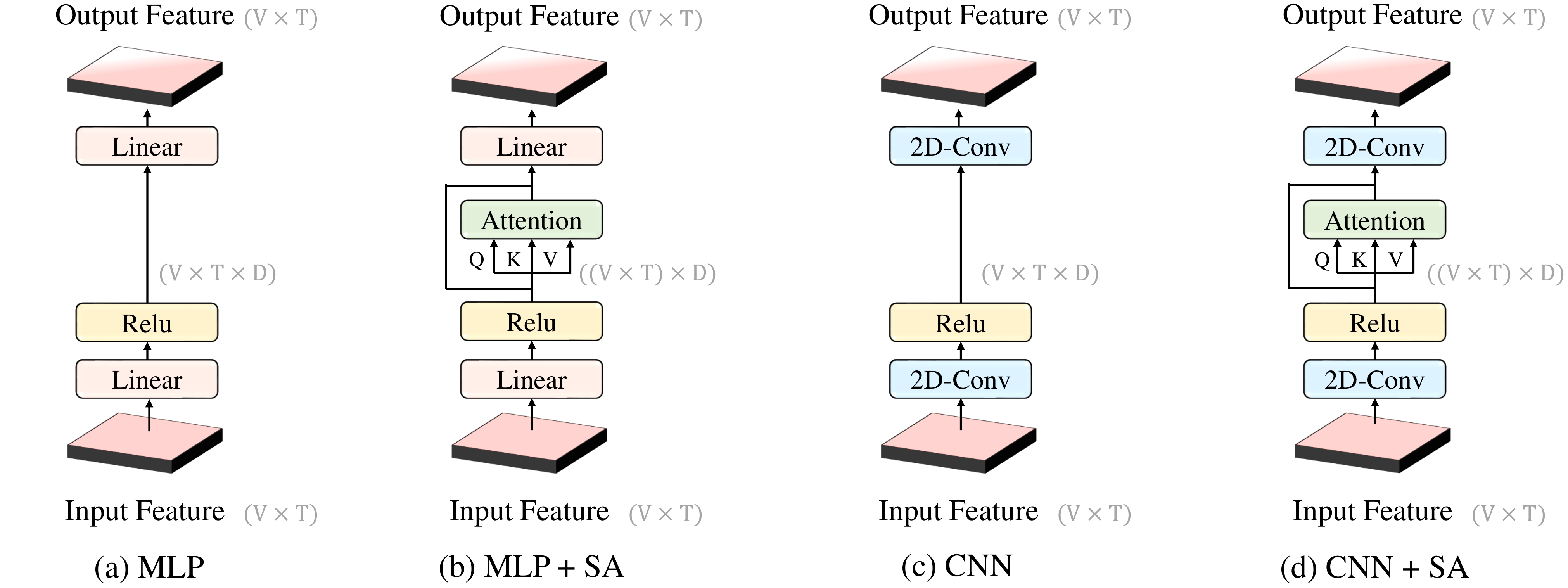}
\caption{\textbf{The structure of the prediction header.} We choose four several popular structures, \ie, ``MLP'', ``CNN'', ``MLP+SA'' and ``CNN+SA''. $V, T, D$ represent the number of visual tokens, the number of textual tokens, and the number of feature channels, respectively.}
\label{fig:prediction}
\end{figure*}

\subsection{Symmetry Axiom}
Symmetry states that if changing the value of two coalitions has the same effect on the output under all values of the other variables, then both coalitions should have an identical interaction score. 

\myparagraph{Proof.} We consider $\mathcal{C}=\{i,j\}, \mathcal{C^{'}}=\{i^{'},j^{'}\}$ fixed. Let us choose $\mathcal{T}\subseteq \mathcal{N}$, and consider the unanimity game. Clearly, $\phi(\mathcal{T} \cup \{[\{i,j\}\})-\phi(\mathcal{T} \cup \{[\{i^{'},j^{'}\}]\})=0, \phi(\mathcal{T}\cup {i})-\phi(\mathcal{T}\cup {i^{'}})=0, \phi(\mathcal{T}\cup {j})-\phi(\mathcal{T}\cup {j^{'}})=0$. That is, for every $\mathcal{T}\subseteq \mathcal{N}$, $\mathcal{C}=\{i,j\}$ and $\mathcal{C^{'}}=\{i^{'},j^{'}\}$ produce the same benefits. Thus, Banzhaf Interaction satisfies Symmetry axiom, \ie, $\mathcal{I}([\mathcal{C}])=\mathcal{I}([\mathcal{C^{'}}])$.

\subsection{Dummy Axiom}
Dummy states that if changing the value of a coalition $[\mathcal{C}]$ has no effect on the output under all values of other variables, then the interaction value of $[\mathcal{C}]$ should be zero. 

\myparagraph{Proof.} We consider $\mathcal{C}=\{i,j\}$ fixed. Let us choose $\mathcal{T}\subseteq \mathcal{N}$, and consider the unanimity game. Clearly, $\phi(\mathcal{S} \cup \{[\mathcal{C}]\})-\phi(\mathcal{S})=0, \sum_{i \in \mathcal{C}} \phi(\mathcal{S}\cup {i})=0$. For every $\mathcal{T}\subseteq \mathcal{N}$, $\mathcal{C}=\{i,j\}$ has no interaction with any player. Thus, Banzhaf Interaction satisfies Dummy axiom, \ie, $\mathcal{I}([\mathcal{C}])=0$.

\subsection{Additivity Axiom}
Additivity states the sum of the interaction scores of the two characteristic functions is equal to the interaction score of the sum of these characteristic functions. 

\myparagraph{Proof.} Let us choose $\mathcal{T}\subseteq \mathcal{N}$, and consider the unanimity game. Clearly, for the characteristic function $\Phi(*)=\phi(*)+\phi^{'}(*)$, $\Phi(\mathcal{T})=\phi(\mathcal{T})+\phi^{'}(\mathcal{T})$. That is, for every $\mathcal{T}\subseteq \mathcal{N}$, the sum of the scores of the two characteristic functions ($\phi(*),\phi^{'}(*)$) is equal to the score of the sum of these characteristic functions $\Phi(*)$. Thus, Banzhaf Interaction satisfies Additivity axiom.

\subsection{Recursivity Axiom}
We hypothesize that the interaction score should depend on the values of $i$ when $j$ is absent, and $j$ when $i$ is absent. And somehow, their interaction should also be taken into account. Specifically,
Recursivity states that if the interaction is positive, then the interaction score of $[\{i,j\}]$ should be greater than simply the sum of individual values. If the interaction is negative, the interaction score of $[\{i,j\}]$ should be less than the sum.

\myparagraph{Proof.} We can rewrite Eq.~\ref{apendix:BI} as $ \mathcal{I}([\mathcal{C}])=\mathcal{B}([\mathcal{C}]|\mathcal{N} \setminus \mathcal{C}\cup \{[\mathcal{\mathcal{C}}]\})-\sum_{i\in \mathcal{C}}\mathcal{B}(i|\mathcal{N} \setminus \mathcal{C}\cup \{i\})$. Clearly, the above formula is equivalent to Recursivity axiom. Thus, Banzhaf Interaction satisfies Recursivity axiom.

\section{Discussions}
\subsection{Banzhaf Interaction Estimator}
Since the calculation of the exact Banzhaf Interaction is an NP-hard problem~\cite{matsui2001np}, existing methods mainly use sampling-based methods~\cite{leech2002computation,bachrach2010approximating} to obtain unbiased estimates. To speed up the computation of Banzhaf Interaction for many data instances, we pre-train a tiny model to learn a mapping from a set of input features to a result using MSE loss. The tiny model consists of a convolutional layer for encoding features, a self-attention module for capturing global interaction, and a convolutional layer for decoding. The tiny model has 64 hidden channels. The input is the similarity matrix of video frames and text tokens, and the output is the estimation of Banzhaf Interaction. 

To explore the impact of the Banzhaf Interaction estimator on our method, we compare the sampling-based method and pre-trained tiny model estimator in Tab.~\ref{tab:ban}. Given the costly training time, the ablation study is based on a subset of MSRVTT dataset (3K videos, each with 20 text descriptions). We find that the pre-trained tiny model maintains the estimation accuracy while avoiding intensive computations. The average training time is reduced from 19.79 seconds per iteration to 3.14 seconds per iteration.

\begin{table}[tb]
\tabfootnotesize
\centering
\setlength{\tabcolsep}{1.5pt}
\begin{tabular}{lccccc}
\toprule[1.25pt]
\multirow{2}{*}{Method} & \multicolumn{4}{c}{{Text-\textgreater{}Video}} & Iteration \\
\cline{2-5}
  & R@1$\uparrow$ & R@5$\uparrow$ & R@10$\uparrow$ & MnR$\downarrow$ & Time$\downarrow$  \\
 \midrule
Baseline & 40.0 & 66.8 & 77.0 & 16.5 & 2.06 s\\
\midrule
w/ Sampling-based method  & 41.5 & \textbf{68.6} & 78.9 & \textbf{15.1} & 19.79 s\\
\rowcolor{aliceblue!60} w/ Tiny estimator  & \textbf{41.8} & 67.5 & \textbf{79.0} & 15.2  & \textbf{3.14} s\\
\bottomrule[1.25pt]
\end{tabular}
\caption{\textbf{Effect of the Banzhaf Interaction Estimator.}  ``$\uparrow$'' denotes that higher is better. ``$\downarrow$'' denotes that lower is better.}
\label{tab:ban}
\end{table}

\subsection{The Structure of the Prediction Header}
Due to the disparity in semantic similarity and interaction index, we design a prediction header to predict the fine-grained relationship $\mathcal{R}_{i,j}$ between the $i_{th}$ video frame and the $j_{th}$ text word. To explore the impact of the structure of the prediction header on our method, we compare four popular structures, \ie, ``MLP'', ``CNN'', ``MLP+SA'' and ``CNN+SA''. Fig.~\ref{fig:prediction} illustrates the structures.

\begin{table}[tb]
\centering
\tabfootnotesize
\setlength{\tabcolsep}{5.2pt}
\begin{tabular}{lccccc}
\toprule[1.25pt]
\multirow{2}{*}{Method} &  \multicolumn{5}{c}{{Text-\textgreater{}Video}}  \\
\cline{2-6}
  & R@1$\uparrow$ & R@5$\uparrow$ & R@10$\uparrow$ & MdR$\downarrow$ & MnR$\downarrow$ \\ \midrule
MLP  & 47.2 & 73.7 & 83.5 & \textbf{2.0} & 12.3\\
CNN  & 47.3 & 73.5 & \textbf{83.7} & \textbf{2.0} & 12.2\\
\midrule
MLP+SA  & 46.6 & 74.0 & \textbf{83.7} & \textbf{2.0} & 12.3\\
\rowcolor{aliceblue!60} CNN+SA   & \textbf{48.6} & \textbf{74.6} & 83.4 & \textbf{2.0} & \textbf{12.0} \\ \bottomrule[1.25pt]
\end{tabular}
\caption{\textbf{Effect of the structure of the prediction header on MSRVTT dataset.} ``SA'' is the self-attention module. ``$\uparrow$'' denotes that higher is better. ``$\downarrow$'' denotes that lower is better.}
\label{tab:prediction head}
\end{table}

\subsection{Self-Distillation}

\begin{table}[tb]
\tabfootnotesize
\centering
\setlength{\tabcolsep}{3.5pt}
\begin{tabular}{lccccc}
\toprule[1.25pt]
\multirow{2}{*}{Method} & \multicolumn{5}{c}{{Text-\textgreater{}Video}}\\
\cline{2-6}
  & R@1$\uparrow$ & R@5$\uparrow$ & R@10$\uparrow$ & MdR$\downarrow$ & MnR$\downarrow$ \\
 \midrule
E-\textgreater{}A & 48.1 & 73.6 & 82.9 & \textbf{2.0} & 11.9\\
E-\textgreater{}O & 48.0 & 74.1 & 83.2 & \textbf{2.0} & \textbf{11.8}\\
A-\textgreater{}O & 48.0 & 73.0 & 83.1 & \textbf{2.0} & 12.0\\
E-\textgreater{}A +\ A-\textgreater{}O & 48.2 & 74.1 & 82.9 & \textbf{2.0} & \textbf{11.8}\\
\rowcolor{aliceblue!60} E-\textgreater{}A\ +\ E-\textgreater{}O & \textbf{48.6} & \textbf{74.6} & \textbf{83.4} & \textbf{2.0} & 12.0\\
\bottomrule[1.25pt]
\end{tabular}
\caption{\textbf{Ablation study about the self-distillation of our method on MSRVTT dataset.} $E,A,O$ denote entity level, action level, and event level, respectively. $-\textgreater{}$ indicates the distillation direction. For example, $E-\textgreater{}A$ indicates the distillation from $E$ to $A$. ``$\uparrow$'' denotes that higher is better. ``$\downarrow$'' denotes that lower is better.}
\label{tab:self}
\end{table}

\begin{table}[tb]
\centering
\tabfootnotesize
\setlength{\tabcolsep}{1.8pt}
\begin{tabular}{ccccc}
\toprule[1.25pt]
{$\mathcal{L}_{I}$ Banzhaf} & {Deep} & {$\mathcal{L}_{D}$ Self} & \multirow{2}{*}{Top1 Acc $\uparrow$} & \multirow{2}{*}{Top5 Acc $\uparrow$}  \\ 
  {Interaction} & {Supervision} & {Distillation} & & \\ \midrule
& &  & 45.2 & 73.1 \\ \midrule
 \scriptsize{\Checkmark} & &  & 45.8 & 73.7 \\
 & \scriptsize{\Checkmark} &  & 46.0 & 74.0 \\
  & \scriptsize{\Checkmark} & \scriptsize{\Checkmark}  & 46.0 & 74.1 \\
\scriptsize{\Checkmark}  & \scriptsize{\Checkmark} & & 46.1 & \textbf{74.2} \\
\rowcolor{aliceblue!60}  \scriptsize{\Checkmark} & \scriptsize{\Checkmark} & \scriptsize{\Checkmark}  & \textbf{46.2} & \textbf{74.2} \\
\bottomrule[1.25pt]
\end{tabular}
\caption{\textbf{Ablation study about the importance of each part on MSRVTT-QA dataset.} ``$\uparrow$'' denotes that higher is better.}
\label{tab:a-vqa}
\end{table}

\textbf{``MLP''} consists of a linear layer with a Relu activation function for encoding features and a linear layer for decoding. The dimension of the hidden channels is 64.
\textbf{``CNN''} consists of a convolutional layer with a Relu activation function for encoding features and a convolutional layer for decoding. The dimension of the hidden channels is 64.
\textbf{``MLP+SA''} consists of a linear layer with a Relu activation function for encoding features, a self-attention module for capturing global interaction, and a linear layer for decoding. The dimension of the hidden channels is 64.
\textbf{``CNN+SA''} consists of a convolutional layer with a Relu activation function for encoding features, a self-attention module for capturing global interaction, and a convolutional layer for decoding. The dimension of the hidden channels is 64.

\begin{table}[tb]
\tabfootnotesize
\centering
\setlength{\tabcolsep}{5.2pt}
\begin{tabular}{lccccc}
\toprule[1.25pt]
\multirow{2}{*}{Method} & \multicolumn{5}{c}{{Text-\textgreater{}Video}}\\
\cline{2-6}
  & R@1$\uparrow$ & R@5$\uparrow$ & R@10$\uparrow$ & MdR$\downarrow$ & MnR$\downarrow$ \\
 \midrule
Baseline & 46.6 & 73.1 & 83.0 & \textbf{2.0} &  13.3\\
\midrule
One level & 47.5 & 73.7 & 83.0 & \textbf{2.0} & 12.0\\
Two levels & 48.1 & 73.6 & 82.9 & \textbf{2.0} & \textbf{11.9}\\
\rowcolor{aliceblue!60} Three levels & \textbf{48.6} & \textbf{74.6} & \textbf{83.4} & \textbf{2.0} & 12.0\\
\bottomrule[1.25pt]
\end{tabular}
\caption{\textbf{Effect of the number of semantic levels (the number of stacked token merge modules) on MSRVTT dataset.} ``$\uparrow$'' denotes that higher is better. ``$\downarrow$'' denotes that lower is better.}
\label{tab:num}
\end{table}

\begin{figure}[tbp]
\centering
\includegraphics[width=.98\linewidth]{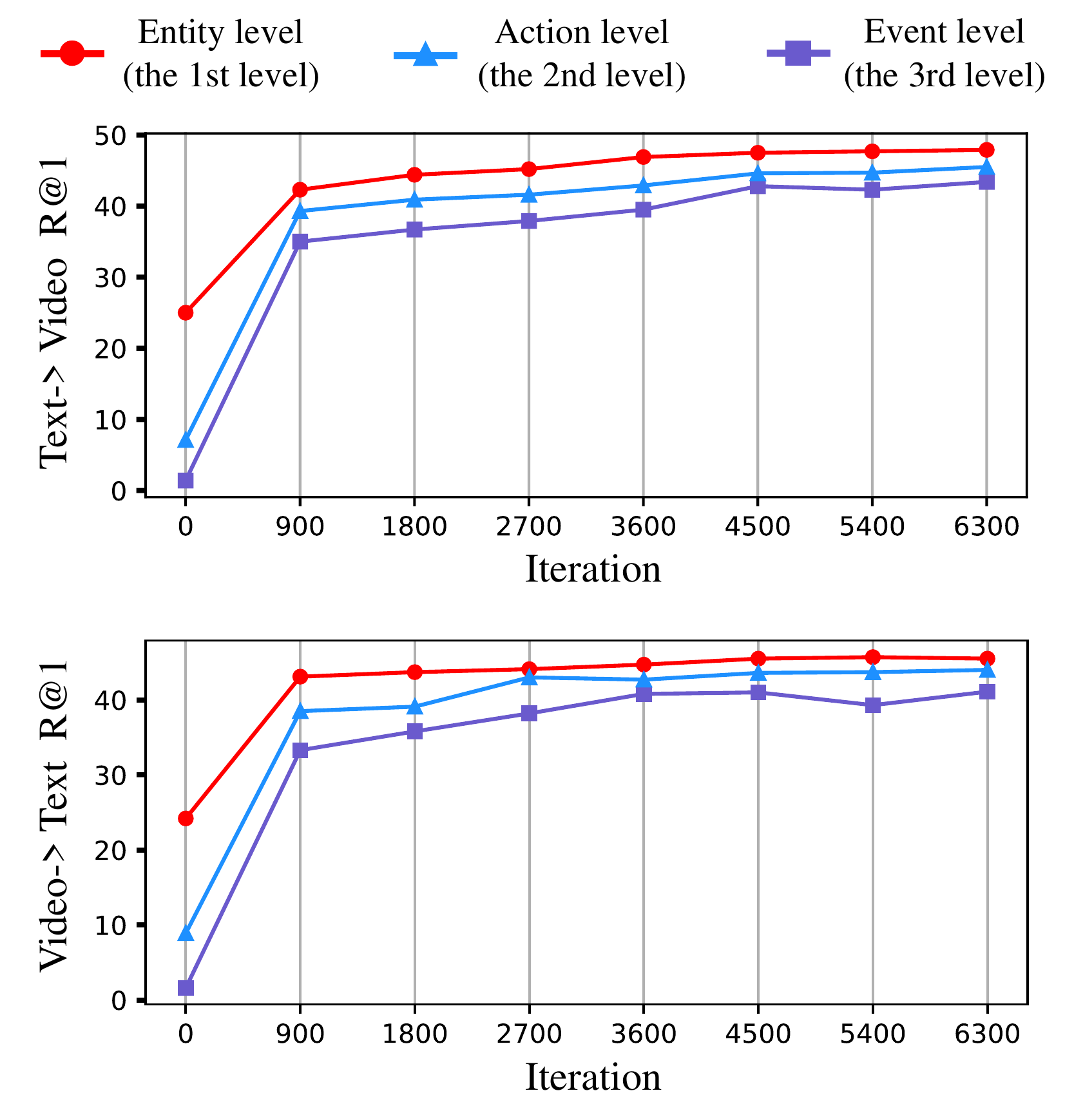}
\caption{\textbf{Performance at each semantic level of text-to-video retrieval and video-to-text retrieval task.}}
\label{fig:self}
\end{figure}

As shown in Tab.~\ref{tab:prediction head}, we find that the combination of CNN and attention (``CNN+SA'') can capture both local and global interaction, so it is beneficial for predicting the fine-grained relationship between video and text. As a result, we adopt ``CNN+SA'' to achieve the best performance.

\begin{figure*}[tbp]
\centering
\includegraphics[width=0.98\linewidth]{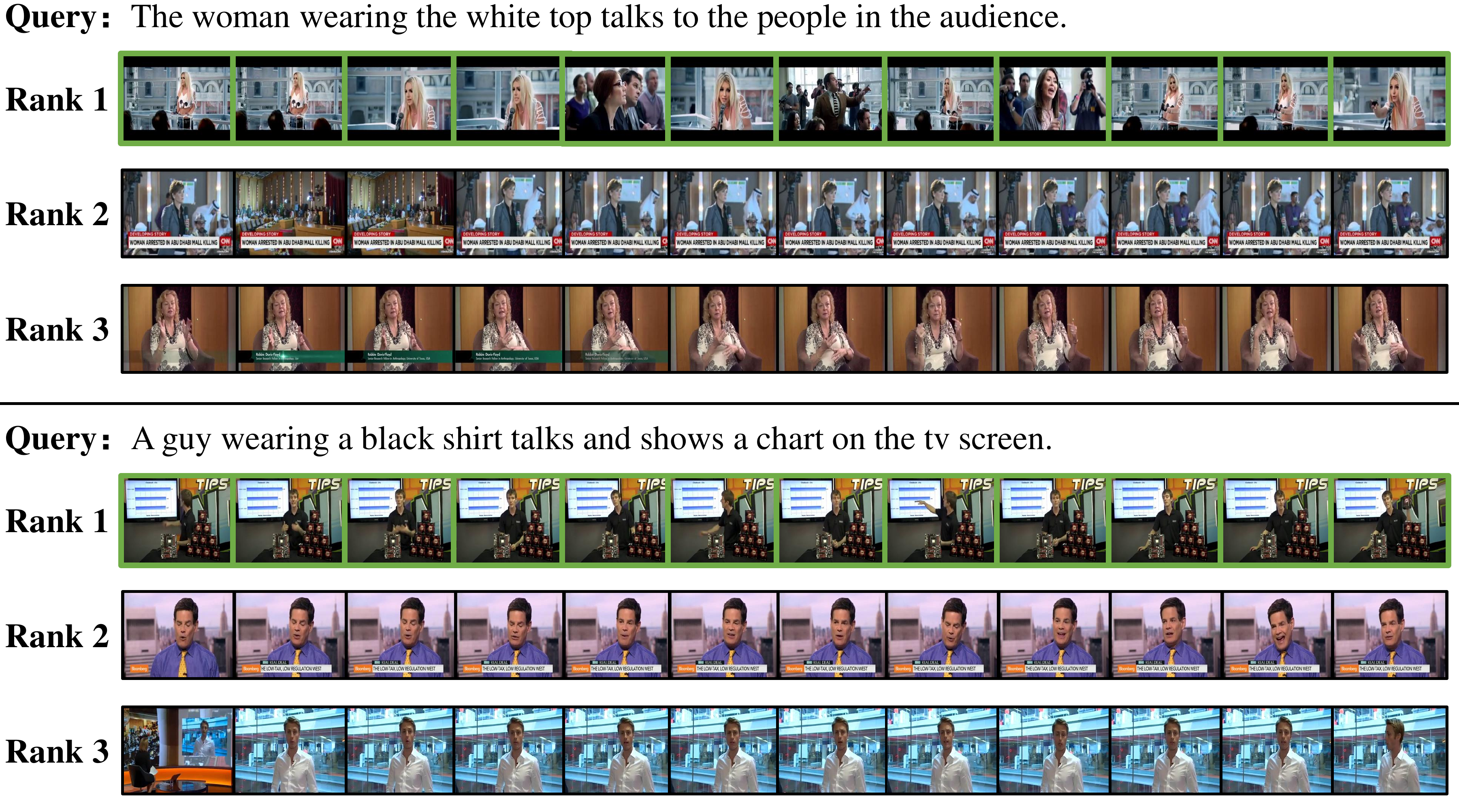}
\caption{\textbf{Visualization of the text-to-video retrieval.} Only the correct videos are highlighted in green.}
\label{fig:retrieval}
\end{figure*}

\begin{figure*}[tbp]
\centering
\includegraphics[width=0.98\linewidth]{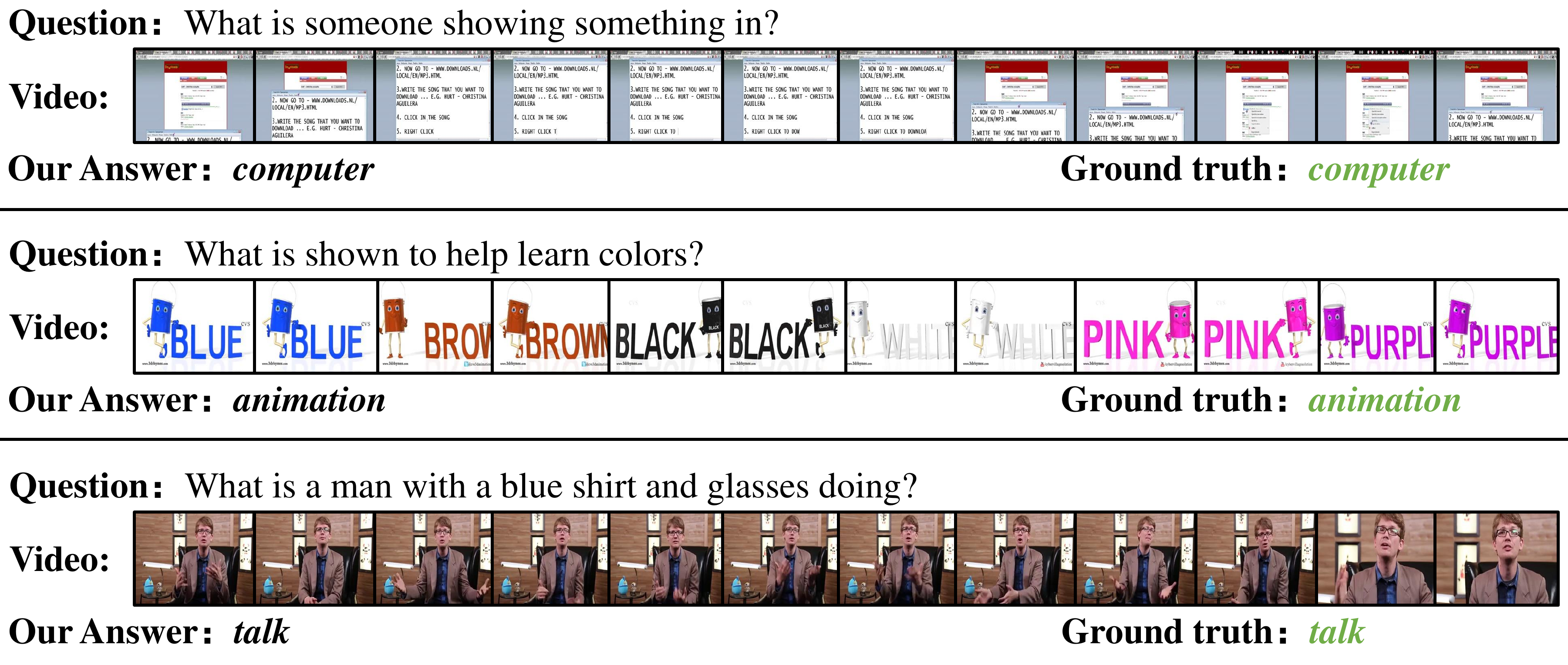}
\caption{\textbf{Visualization of the video-question answering.}}
\label{fig:vqa}
\end{figure*}

\begin{figure*}[tbp]
\centering
\includegraphics[width=1\linewidth]{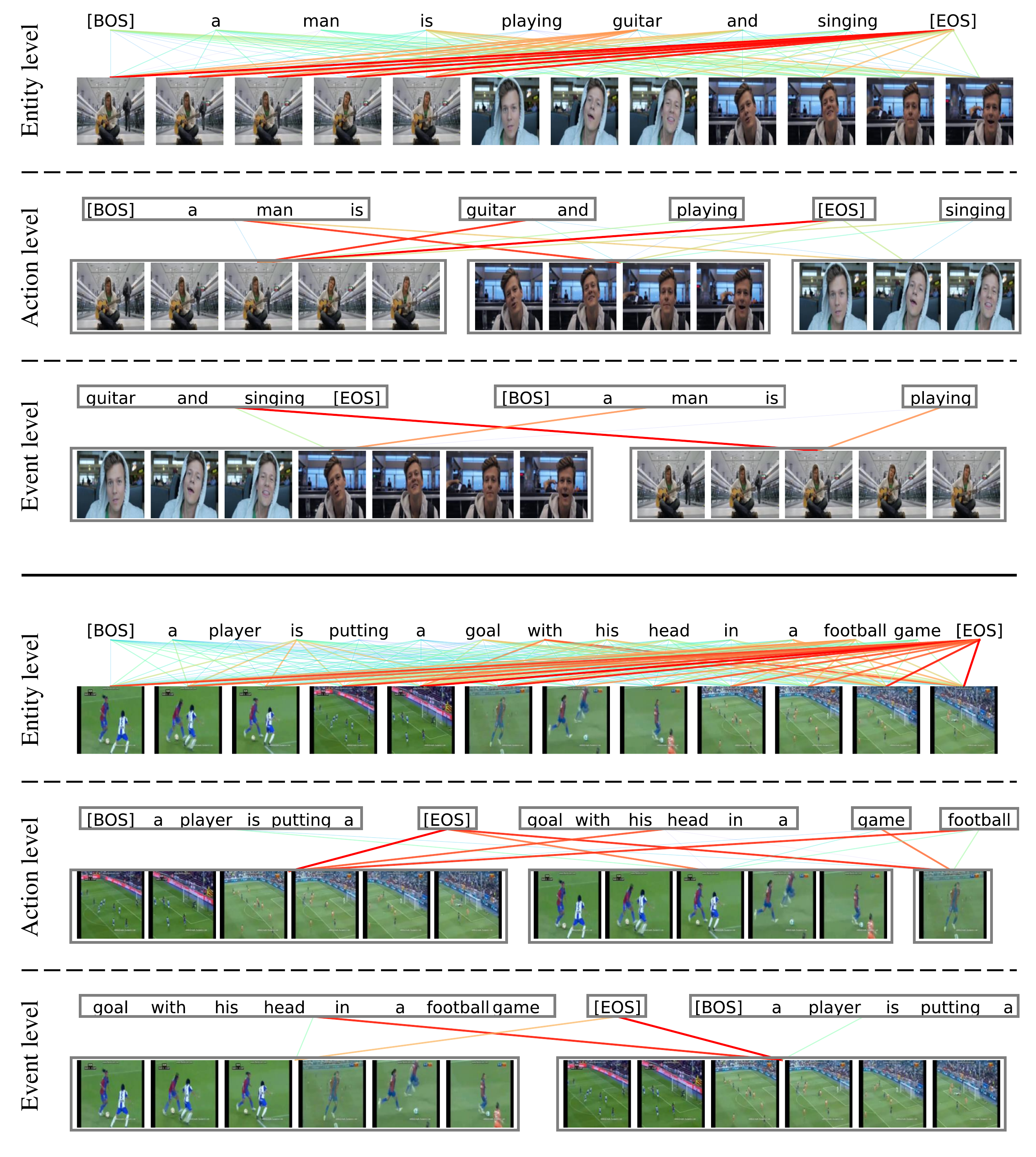}
\caption{\textbf{Visualization of the hierarchical interaction.} Here, the degree of confidence from high to low is represented by red, orange, green and blue lines, respectively. Entity-level interactions demonstrate the semantic correlation between frames and words. Action-level interactions indicate the semantic correlation between clips and phrases. Event-level interactions show the semantic correlation between segments and paragraphs.}
\label{fig:hierarchical_0}
\end{figure*}

\begin{figure*}[tbp]
\centering
\includegraphics[width=1.\linewidth]{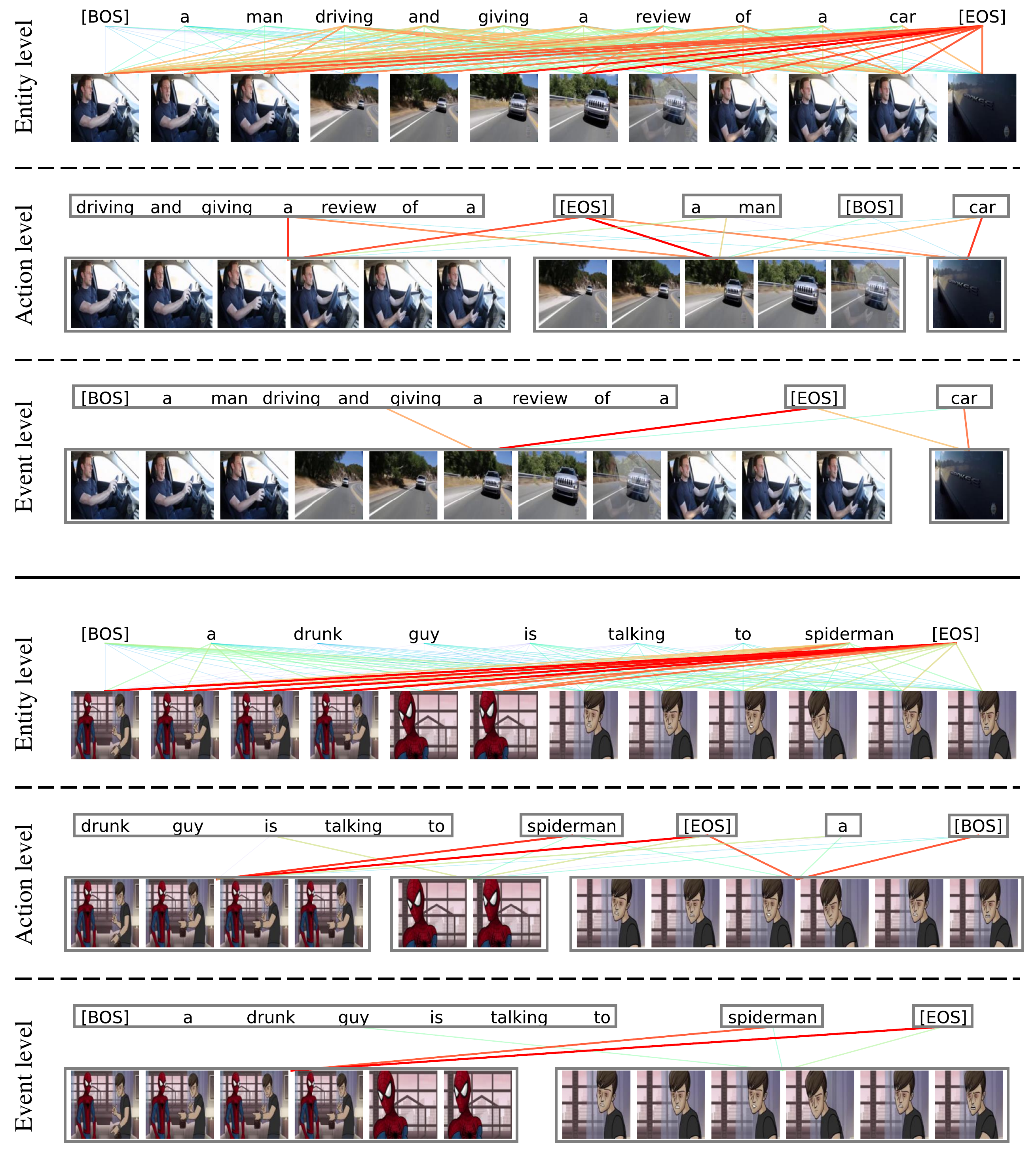}
\caption{\textbf{Visualization of the hierarchical interaction.} Here, the degree of confidence from high to low is represented by red, orange, green and blue lines, respectively. Entity-level interactions demonstrate the semantic correlation between frames and words. Action-level interactions indicate the semantic correlation between clips and phrases. Event-level interactions show the semantic correlation between segments and paragraphs.}
\label{fig:hierarchical_1}
\end{figure*}

\begin{figure*}[tbp]
\centering
\includegraphics[width=1\linewidth]{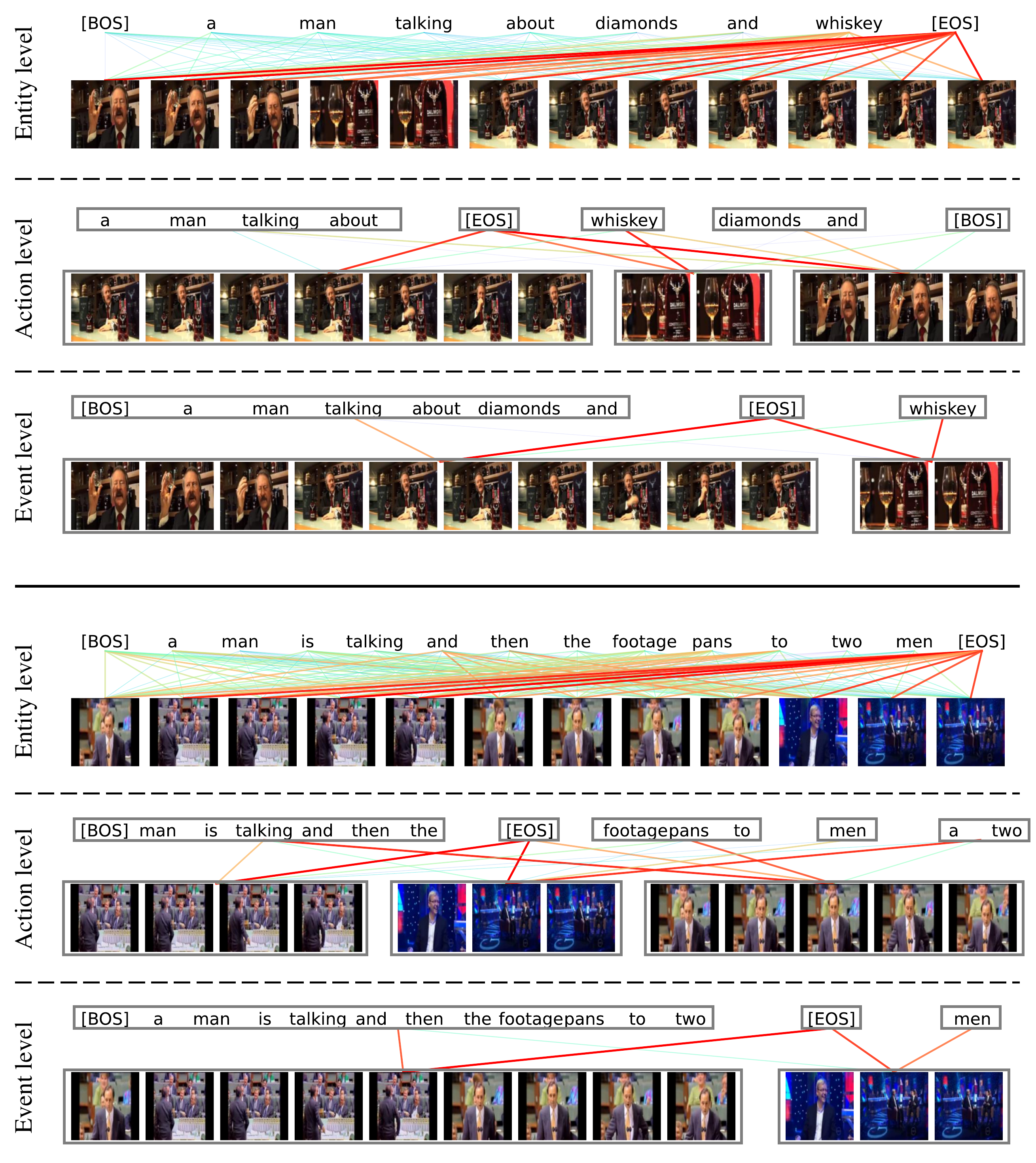}
\caption{\textbf{Visualization of the hierarchical interaction.} Here, the degree of confidence from high to low is represented by red, orange, green and blue lines, respectively. Entity-level interactions demonstrate the semantic correlation between frames and words. Action-level interactions indicate the semantic correlation between clips and phrases. Event-level interactions show the semantic correlation between segments and paragraphs.}
\label{fig:hierarchical_2}
\end{figure*}

Fig.~\ref{fig:self} shows the performance of each semantic level. We find that the entity level converges first in the training process. This is because higher-level semantic features are merged from lower-level semantic features. When lower-level semantic features do not converge, it is difficult for higher-level semantic features to learn semantic information. Based on this observation, we propose using lower-level semantic features to guide the learning of higher-level semantic features. Thus, we distill the entity-level similarity to the other two semantic levels.

To illustrate the impact of the self-distillation of our method, we conduct ablation experiments on MSRVTT dataset in Tab.~\ref{tab:self}. As we can see, self-distillation improves the generalization ability. Distilling from the entity level to the other two semantic levels achieves the best results. As a result, we distill the entity-level similarity to the other two semantic levels as default in practice.

\subsection{Ablation for Video-Question Answering Task}
To illustrate the importance of each part of our method for the video-question answering, we conduct ablation experiments on MSRVTT-QA dataset in Tab.~\ref{tab:a-vqa}. As we can see, Banzhaf Interaction boosts the baseline with the improvement up to 0.6\% at Top1 accuracy. Moreover, deep supervision and self-distillation significantly improve the generalization ability. Self-distillation provides limited improvement for video-question answering compared to text-video retrieval. This is because reasoning relies primarily on high-level semantic features. Therefore, it is difficult for low-level semantic features to guide high-level semantic features. Our full model achieves the best performance and outperforms the baseline by 1.0\% at Top1 accuracy.

\subsection{The Number of Semantic Levels}
To efficiently generate coalitions among game players, we cluster the original visual (textual) tokens and compute the Banzhaf Interaction between the merged tokens. By stacking token merge modules, we get cross-modal interaction efficiently at different semantic levels.

To explore the impact of the number of semantic levels on our method, we conduct ablation experiments on MSRVTT dataset in Tab.~\ref{tab:num}. We find that the performance of the model increases with the number of semantic levels. These results indicate that stacking more token merge modules can provide more coalitions, which enables the model to learn more diverse semantic interaction information. We make a trade-off between the number of semantic levels and computation cost and set the number of semantic levels to 3 in practice.

\subsection{Limitations of our Work}
The cross-modal contrastive approach typically exploits the coarse-grained labels of video-text pairs to learn a global semantic interaction. To move a step further, we model video-text as game players with multivariate cooperative game theory to handle the uncertainty during fine-grained semantic interaction with diverse granularity, flexible combination, and vague intensity. Therefore, our method inevitably requires more training time costs. Although our method takes less time than TS2-Net~\cite{liu2022ts2} during the inference stage (see Tab.~5 in the main paper), more effort could be paid to obtain an efficient structure in the future.

\section{Visualizations}
\subsection{Text-to-Video Retrieval}
We show two retrieval examples from the MSR-VTT testing set for text-to-video retrieval in Fig.~\ref{fig:retrieval}. As shown in Fig.~\ref{fig:retrieval}, our method successfully retrieves the ground-truth video. These results demonstrate that our method can align video and text effectively.

\subsection{Video-Question Answering}
We show the visualization of the video-question answering in Fig.~\ref{fig:vqa}. As shown in Fig.~\ref{fig:vqa}, our method succeeds in getting the ground-truth answer. These results demonstrate that our method can deal with cross-modal inference task effectively.

\subsection{Hierarchical Interaction}
To better understand the proposed method, we show the visualization of the hierarchical interaction in Fig.~\ref{fig:hierarchical_0}, Fig.~\ref{fig:hierarchical_1} and Fig.~\ref{fig:hierarchical_2}. This experiment shows that our Hierarchical Banzhaf Interaction (HBI) can effectively handle fine-grained semantic interaction with diverse granularity, flexible combination, and vague intensity. More encouragingly, the visualization illustrates that the proposed method can be used as a tool for visualizing the cross-modal interaction and help us understand the cross-modal model.

\end{document}